\documentclass{article} 
\usepackage{iclr2027_conference,times}

\usepackage{amsmath,amsfonts,bm}

\def\eqref#1{equation~\ref{#1}}

\def\1{\bm{1}}

\DeclareMathAlphabet{\mathsfit}{\encodingdefault}{\sfdefault}{m}{sl}
\SetMathAlphabet{\mathsfit}{bold}{\encodingdefault}{\sfdefault}{bx}{n}

\usepackage{wrapfig}
\usepackage{longtable}
\usepackage{float}
\usepackage{tikz}
\usepackage{placeins}
\usepackage{graphicx}
\usepackage{hyperref}
\usepackage{url}
\usepackage{multirow}
\usepackage{amsmath, amssymb, amsthm}
\usepackage{booktabs}
\usepackage{adjustbox}
\usepackage{capt-of}
\usepackage{pgfplots}
\pgfplotsset{compat=1.18}
\title{GeoPID: Decomposing and Steering Visual Information in Vision-Language Models}

\author{
  \textbf{Seulgi Kim}$^{1}$\thanks{Work done as an intern at Amazon.} \qquad \textbf{Zhixiong Zhang}$^2$ \qquad \textbf{Xinwei Zhang}$^2$ \qquad \textbf{Jie Ling}$^2$ \qquad \textbf{Ronn Shaw}$^2$ \\
  $^1$Georgia Institute of Technology \qquad $^2$Amazon
}

\iclrfinalcopy 
\begin{document}

\maketitle


\begin{abstract}
While recent vision-language models (VLMs) have shown outstanding performance across diverse applications, they tend to under-use visual information and over-rely on textual context. In this work, we propose \textsc{GeoPID}, a training-free framework that analyzes multimodal information within VLMs from a geometric perspective. \textsc{GeoPID} decomposes information into Redundant, Modality-Unique, and Synergistic components through the geometric relationships between visual and textual representation subspaces. Through an extensive analysis across 22 VLMs and 14 benchmarks, we confirm that correct predictions exhibit stronger vision-unique components when questions strongly require visual grounding. Building on this geometric analysis, we introduce a targeted intervention technique that selectively amplifies visual representations along the vision-unique subspace during inference. As a result, visual grounding capabilities were enhanced without any additional model parameter updates, achieving an average relative accuracy gain of 7.63\%.
\end{abstract}
\section{Introduction}
\begin{figure}[ht!]
\centering
\includegraphics[scale =0.53]{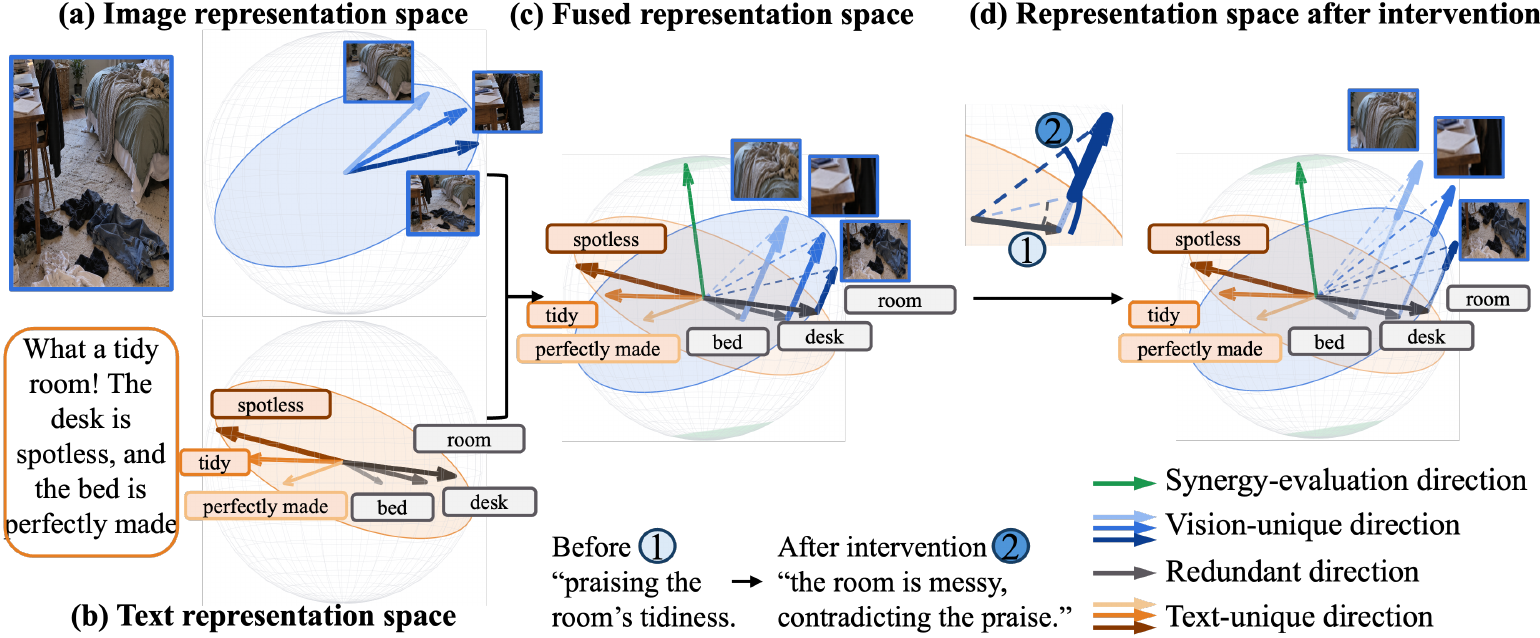}
\caption{In this figure, we take a visual and textual input and project them into their own representation space (a-b), where vectors represent features that can be extracted from each input modality. (c): The fused representation space reveals geometric relationships and contrasting modality-specific content: text-unique features (\textcolor{orange}{orange}) praise the room's tidiness, while vision-unique features (\textcolor{blue}{blue}) depict the disorder. The \textcolor{green}{green} direction denotes the complementary geometry in which synergy is evaluated. (d): Our approach selectively amplifies along the vision-unique subspace while preserving its orthogonal complement.}
\label{fig:teaser}
\vspace{-0.5cm}
\end{figure}

Recent advances in vision-language models (VLMs) enable joint grounding of visual and linguistic information, resulting in state-of-the-art performance across a wide range of practical applications. This includes visual question answering~\cite{zhang2024pmc}, action understanding~\citep{kaviani2025hierarchical, kim2025multi}, medical image analysis~\citep{li2023llava}, and judging the quality of generated visual data~\citep{lin2024evaluating}. A core motivation behind VLMs is to combine visual and linguistic information, which can include both shared context and modality-specific details. Ideally, a model should use the visual information required by the question and combine it with textual instructions to produce an image-grounded answer. For example, in Figure~\ref{fig:teaser}(c), both modalities establish the shared context of a bedroom, where the text expresses explicit praise while the image depicts clear disorder. Jointly processing these inputs reveals a mismatch between what is said and what is shown, which is essential for capturing the speaker's sarcastic intent as shown in Figure~\ref{fig:teaser}(d-2). However, VLMs often under-use visual information and over-rely on textual context~\citep{leng2024mitigating}. For example, in Figure~\ref{fig:teaser}(d-1), the model interprets the text as genuine praise, missing the visual evidence needed to recognize the contradiction. This raises specific questions: do VLMs use visual information beyond what is already available from the text, and how is this information organized within their internal representations?

Understanding how to leverage visual information, therefore requires distinguishing information shared with text, information unique to vision, and information available only through their combination. Partial Information Decomposition (PID)~\citep{williams2010nonnegative} provides a framework for making these distinctions by decomposing the information that multiple modalities provide into three types: \textbf{redundant information (R)}, which is available in both vision and language; \textbf{unique information (U)}, which is provided exclusively by one modality; and \textbf{synergistic information (S)}, which becomes available only when the two modalities are considered jointly. Figure~\ref{fig:teaser} illustrates these distinctions. Both modalities provide shared content (e.g., `room', `desk', `bed') but they also contain contrasting unique content: in Figure~\ref{fig:teaser}(b), the text characterizes the scene as tidy, spotless, and perfectly made, whereas the image in Figure~\ref{fig:teaser}(a,c) provides image-specific details showing scattered clothes, a cluttered desk, and rumpled bedding. Recognizing the mismatch between the textual praise and the visual scene illustrates the role of synergistic information, supporting a sarcastic interpretation that neither modality establishes alone. Shared information alone may not be sufficient (Figure~\ref{fig:teaser}(d-1)`Before: praising the room's tidiness') as it misses the cross-modal relationship. The goal is therefore to distinguish these information components in VLMs, and use these distinctions to determine which visual components to selectively amplify to improve predictions.

Translating these distinctions into selective amplification presents two challenges. First, existing approaches quantify PID components using separate estimators, often based on predictions~\citep{fang2026towards,liang2023multimodal}, but numerical PID values alone do not identify where shared and modality-specific information resides. Figure~\ref{fig:teaser} illustrates this structure using 3D spheres to represent hidden representation spaces, with arrows corresponding to distinct features. Figure~\ref{fig:teaser}(c) displays their geometric relationships in a fused space. Gray arrows denote shared features, orange arrows represent text-unique contents, and blue arrows isolate vision-unique details. Identifying these specific subspaces is important because it provides a target for selectively amplifying the vision-unique directions, as in Figure~\ref{fig:teaser}(d). Second, improvements informed by multimodal information decomposition have primarily required costly additional training that is not practical in many application settings. Existing training-free methods for improving visual grounding at inference time instead rely on heuristic signals. This includes contrasting with distorted images~\citep{leng2024mitigating}, generating captions~\citep{li2026geass}, or steering vectors~\citep{yin2026dynamic}, without explicitly decomposing shared, unique, and synergistic information to identify the intervention target.

To address these limitations, we propose \textsc{GeoPID}, a training-free framework for understanding and steering visual information use in VLMs. Our central idea is to characterize multimodal information through the geometric relationships between visual and textual representation subspaces. Intuitively, a modality's representations can span directions that align with the other modality as well as directions that remain distinct. Based on this geometric perspective, we propose representation-based measures of PID components. To characterize geometric redundancy (\textbf{R}), we use the principal angles between the visual and textual representation subspaces to measure the extent to which the two modalities span shared directions. The unique visual and textual components, denoted by $\mathbf{U_v}$ and $\mathbf{U_t}$, are associated with modality-specific directions orthogonal to the other modality's estimated subspace. For synergy (\textbf{S}), we use the geometry outside the principal span of the combined visual and textual representations as the space in which to measure answer-related information. However, information in this space does not necessarily come from combining corresponding visual and textual inputs. We therefore project decision representations onto this complementary space and define our synergy measure as the difference in mutual information between these projected representations and answer labels under matched and mismatched image--text pairs. By connecting information decomposition to representation geometry, \textsc{GeoPID} provides both an analytical description of visual information and a concrete subspace target for intervention.

We apply \textsc{GeoPID} to $14$ vision-language benchmarks across $22$ models and identify three patterns. First, shared geometric structure develops across early and intermediate layers, while models exhibit different modality-specific profiles in later layers. This indicates that shared structure alone does not establish whether visual information is effectively used. Second, the information components vary with what is needed to answer each question: visual-grounding questions exhibit the highest $\mathbf{U_v}$, while questions requiring visual grounding and the combination of text and image exhibit higher \textbf{S}. Third, for questions requiring vision information, correct answers exhibit stronger vision-unique components in the representation than incorrect answers. This suggests that strengthening representation components along vision-unique directions may help the model answer visually demanding questions. We therefore test a training-free intervention that selectively amplifies vision-token representation components along the identified vision-unique subspace. As illustrated in Figure~\ref{fig:teaser}(c), this intervention strengthens image-specific components while preserving the remaining representation components, allowing subsequent layers to integrate the amplified visual content into the prediction. Our intervention yields an average relative accuracy gain of 7.63\%, with qualitative examples further illustrating improved visual grounding in open-ended generation. Our contributions are as follows: 
\begin{enumerate}
    \item We present a geometric PID framework that characterizes shared and modality-specific representation subspaces and connects their structure to target-related information in VLMs. 

    \item We analyze how geometric structure and associated information evolve across layers and vary with the information required by a question, providing insight into visual information use, and motivating a vision-unique intervention target. 

    \item We develop a training-free intervention that selectively amplifies representation components along the identified vision-unique directions and evaluate its effectiveness across $14$ benchmarks and $22$ VLMs.
\end{enumerate}

\noindent \textbf{Related Works.} Partial Information Decomposition (PID) separates multimodal contributions into redundant, unique, and synergistic components~\citep{williams2010nonnegative}. While recent works apply PID to VLMs for diagnosis and fine-tuning, they typically rely on separate estimators or require costly retraining~\citep{liang2023multimodal, fang2026towards, wu2026vision}. We instead compute these components directly from representation geometry, advancing beyond overall matrix rank measures~\citep{kim2026countering, chaudhuri2025closer} to explicitly distinguish shared from modality-specific subspaces. Furthermore, existing training-free interventions mitigate visual hallucinations through heuristic methods~\citep{leng2024mitigating, liu2024reducing}. \textsc{GeoPID} differs by explicitly utilizing this geometric PID to identify and selectively amplify the vision-unique subspace during inference, providing a theoretically grounded target rather than relying on heuristics. A comprehensive review of related literature is provided in Appendix~\ref{lab:related-works}.
\providecommand{\acc}{\operatorname{acc}}
\providecommand{\Irr}{\mathcal{I}_{2}}
\providecommand{\Guv}{G_{U_V}}
\providecommand{\Gut}{G_{U_t}}

\section{Methodology}
\label{sec:method}
In this section, we address two complementary components. First, we analyze the geometry formed by vision and text tokens in the hidden representation and define four geometric PID components corresponding to redundancy, modality-specific uniqueness, and synergy. Second, we use the resulting vision-unique subspace as an intervention target and selectively amplify visual representation components along that subspace during inference.

\subsection{Characterizing Multimodal Information}
\label{subsec:pid_diagnosis}
Given a reference batch $\mathcal{B}$ containing $N_{\mathcal{B}}$
image-text samples, let $V_i\in\mathbb{R}^{n_{i}^{v}\times d}, T_i\in\mathbb{R}^{n_{i}^{t}\times d}$
denote the matrices obtained by stacking the vision- and text- token representations of sample $i$, respectively. Here, $n_{i}^{v}, n_{i}^{t}$ denote the number of vision and text tokens in sample $i$, and $d$ is the hidden dimension. Let $h_i\in\mathbb{R}^{d}$ denote the representation of the last input token, and stacking these representations gives $H = \begin{bmatrix} h_1 \dots h_{N_{\mathcal{B}}} \end{bmatrix}^T \in\mathbb{R}^{N_{\mathcal{B}}\times d}$.

We start from the observation that vision- and text- representations do not necessarily occupy the same directions, even when they reside in a shared $d$-dimensional hidden space. A direction refers to a feature axis within this space. As illustrated by the arrows in Figure~\ref{fig:teaser} (c), some directions may be expressed by both modalities, whereas others are predominantly occupied by only one modality. Based on this structure, we define four geometric bases to formalize these multi-modal relationships:
\begin{itemize}
    \item $G_R$: The geometry shared by vision and text, as shown in the gray directions in Figure~\ref{fig:teaser}(c).
    \item $G_{U_V}$: The vision geometry orthogonal to the estimated text subspace. This is illustrated as the blue directions in Figure~\ref{fig:teaser} (c).
    \item $G_{U_t}$: The text geometry orthogonal to the estimated vision subspace. This is illustrated as the orange directions in Figure~\ref{fig:teaser} (c).
    \item $G_S$: The geometry outside the principal span jointly estimated from the vision and text representations. This is illustrated as the green directions in Figure~\ref{fig:teaser} (c).
\end{itemize}

We estimate geometric PID in two stages. First, using only the vision and text token representations, we construct modality subspaces and identify the shared geometry ($G_R$), modality-unique geometries ($G_{U_V}$, $G_{U_t}$), and the geometry in which synergy is evaluated ($G_S$). Second, for offline diagnosis, we measure the target information $Y$ associated with these geometries using matrix-based R\'enyi mutual information, yielding the geometric PID components: $R$, $U_V$, $U_t$, and $S$.

\paragraph{Batch-wise Modality Geometry Estimation.}
Because the number of tokens may differ across samples, we normalize each sample's contribution by its token count before estimating a batch-level covariance matrix. This prevents samples with longer token sequences from disproportionately determining the estimated geometry.

Let $V_{i,p},T_{i,p}\in\mathbb{R}^{d}$ denote the representations corresponding to the $p$-th rows of $V_i$ and $T_i$, respectively, written as column vectors. We compute the vision and text means as
$\mu_v = \frac{1}{N_{\mathcal{B}}} \sum_{i\in\mathcal{B}} \frac{1}{n_i^v} \sum_{p=1}^{n_i^v} V_{i,p}, \mu_t = \frac{1}{N_{\mathcal{B}}} \sum_{i\in\mathcal{B}} \frac{1}{n_i^t} \sum_{p=1}^{n_i^t} T_{i,p}$.
The corresponding covariance matrices are 
$C_v = \frac{1}{N_{\mathcal{B}}} \sum_{i\in\mathcal{B}} \frac{V_i^\top V_i}{n_i^v} - \mu_v\mu_v^\top, C_t = \frac{1}{N_{\mathcal{B}}} \sum_{i\in\mathcal{B}} \frac{T_i^\top T_i}{n_i^t} - \mu_t\mu_t^\top$.
Dividing by $n_i^v$ or $n_i^t$ averages the contributions within each sample, while dividing by $N_{\mathcal{B}}$ averages across samples in the reference batch. Similarly, to construct the combined marginal geometry used to define $G_S$, we apply the same procedure to the vision and text representations concatenated within each sample:
$\mu_{VT} = \frac{1}{N_{\mathcal{B}}} \sum_{i\in\mathcal{B}} \frac{\sum_{p=1}^{n_i^v} V_{i,p} + \sum_{p=1}^{n_i^t} T_{i,p}}{n_i^v+n_i^t}, C_{VT} = \frac{1}{N_{\mathcal{B}}} \sum_{i\in\mathcal{B}} \frac{ V_i^\top V_i+T_i^\top T_i}{n_i^v+n_i^t} - \mu_{VT}\mu_{VT}^\top$.
A single set of covariance matrices $(C_v,C_t,C_{VT})$ is estimated from the reference batch and shared across samples. The leading eigenspaces of $C_v$ and $C_t$ define the principal modality bases used to construct $G_R$, $G_{U_V}$, and $G_{U_t}$. The leading eigenspace of $C_{VT}$ defines the combined marginal span whose orthogonal complement specifies $G_S$.

\paragraph{Shared geometry.}
We first obtain principal bases for the vision and text representations using the batch-wise PCA procedure defined above, where we visualize these subspaces in Appendix~\ref{app:md-subspaces}. Each sample contributes equally regardless of its number of tokens:
\begin{equation}
B_v = \operatorname{PCA}_{r_b} \left({V_i}\right), \qquad B_t = \operatorname{PCA}_{r_b} \left({T_i}\right), \qquad i\in\mathcal{B} \label{eq:modality_bases} 
\end{equation} 
where $\operatorname{PCA}_{r_b}$ denotes the leading $r_b$ principal directions obtained using the sample-normalized covariance estimator defined above. Here, we set $r_b=64$. (Design choice is in Appendix~\ref{app:md-rank-r}). We measure the alignment between these modality subspaces using their principal angles:
\begin{equation}
B_v^\top B_t
=
U\operatorname{diag}(c_1,\ldots,c_{r_b})W^\top,
\end{equation}
where $c_j\in[0,1]$ is the cosine of the $j$-th principal angle. Larger values of $c_j$ indicate stronger alignment between the corresponding vision and text directions. Let $U_{c>\tau}$ contain the columns of $U$ whose principal-angle cosines exceed the alignment threshold $\tau=0.7$. Since we aim to contrast the text-aligned and text-orthogonal directions of the vision geometry, we represent the shared geometry using the vision-side principal directions aligned with the text subspace:
\begin{equation}
G_R
=
B_vU_{c>\tau}.
\end{equation}

\paragraph{Modality-unique geometries.}
We project each modality's representations onto the
orthogonal complement of the other modality's estimated
subspace. For each sample, we form
\begin{equation}
V_{\perp t,i}=V_i(I-B_tB_t^\top),
\qquad
T_{\perp v,i}=T_i(I-B_vB_v^\top).
\end{equation}
We then apply the same batch-wise PCA with rank cap $r_u=32$:
\begin{equation}
G_{U_V}
=
\operatorname{PCA}_{r_u}
\left(\{V_{\perp t,i}\}_{i\in\mathcal{B}}\right),
\qquad
G_{U_t}
=
\operatorname{PCA}_{r_u}
\left(\{T_{\perp v,i}\}_{i\in\mathcal{B}}\right).
\label{eq:unique_geometry}
\end{equation}
Ablation study of $r_u$ is in Appendix~\ref{app:md-rank-r}. Thus, $G_{U_V}$ contains the principal vision directions
orthogonal to the estimated text subspace, while
$G_{U_t}$ contains the principal text directions
orthogonal to the estimated vision subspace. This orthogonality is further explained in Appendix~\ref{app:md-subspaces}.

\paragraph{Geometry for synergy.}
For each sample, we concatenate its vision- and text-token matrices and apply the same batch-wise PCA procedure to the combined token representations and retain their leading $r_s=128$ principal directions (The design choice of $r_s$ is in Appendix~\ref{app:md-rank-r}):
\begin{equation}
B_{VT}
=
\operatorname{PCA}_{r_s}
\left(\begin{bmatrix} V_i\\
T_i \end{bmatrix}\right), \qquad {i\in\mathcal{B}}
\label{eq:synergy_pca}
\end{equation}
We define $G_S$ as an orthonormal basis for the complement of this combined marginal span:
\begin{equation}
\operatorname{span}(G_S)
=
\operatorname{span}(B_{VT})^\perp,
\qquad
G_S^\top G_S=I.
\label{eq:synergy_geometry}
\end{equation}

Consequently, $HG_S$ represents the decision components lying outside the principal directions expressed by the combined vision and text token representations. In implementation, this operation can equivalently be computed as $H\left(I-B_{VT}B_{VT}^\top\right)$, without explicitly materializing a basis for the full orthogonal complement. This implementation process is detailed in Appendix~\ref{app:md-subspaces}.

Importantly, information located in $G_S$ is not automatically regarded as synergy. $G_S$ specifies only the complementary region in which correspondence dependence is evaluated. We operationalize synergy in this region through a contrast between correctly matched and mismatched image--text conditions.

\paragraph{Geometric PID measurement.}
Following prior work on matrix-based information measurement~\citep{giraldo2014measures,yu2019multivariate,yu2018multivariate}, we use matrix-based R\'enyi-2 mutual information, denoted by $I_2$, to measure the dependence between the decision representations ($H$) projected onto each geometry and the target $Y$. For each projected representation matrix $Z$, we normalize its nonzero rows to unit length and construct the cosine Gram matrix $K_Z=\hat Z\hat Z^\top$. For the target, we use the label kernel $K_Y[i,j]=\mathbf{1}[y_i=y_j]$~\citep{wickstrom2019information}, which assigns a value of one to samples with the same target label and zero otherwise.

We first normalize each Gram matrix $K_Z, K_Y$ to have unit trace: $\widehat{K}_Z=\frac{K_Z}{\operatorname{tr}(K_Z)}, \widehat{K}_Y=\frac{K_Y}{\operatorname{tr}(K_Y)}$. Then, we combine the representation similarities and target-label similarities through their element-wise product and normalize the result by $\widehat{K}_{ZY} = \frac{\widehat{K}_Z\circ \widehat{K}_Y} {\operatorname{tr}(\widehat{K}_Z\circ \widehat{K}_Y)}$, where $\circ$ denotes the Hadamard product. Here, because the target kernel is zero for samples with different labels, this operation retains representation similarities only between samples sharing the same target. Following the matrix-based R\'enyi formulation~\citep{giraldo2014measures,yu2019multivariate,yu2018multivariate}, we compute mutual information from the two marginal matrices and their joint matrix as
\begin{equation}
I_2(Z;Y)
=
\mathcal{S}_2(\widehat{K}_Z)
+
\mathcal{S}_2(\widehat{K}_Y)
-
\mathcal{S}_2(\widehat{K}_{ZY}),
\label{eq:renyi_mi}
\end{equation}
where $\mathcal{S}_2(\widehat{K})=-\log\operatorname{tr}(\widehat{K}^2)$ is the matrix-based R\'enyi-2 entropy~\citep{giraldo2014measures,yu2019multivariate,yu2018multivariate}. We compute this directly from kernel entries
without an eigendecomposition, as detailed in Appendix~\ref{app:md-info-proofs}. With Eq.~\ref{eq:renyi_mi}, we define geometric redundancy and modality-specific
uniqueness as
\begin{equation}
R = I_2(HG_R;Y),
\qquad
U_V = I_2(HG_{U_V};Y),
\qquad
U_t = I_2(HG_{U_t};Y).
\label{eq:unique}
\end{equation}
Here, $R$ measures target-related information along the vision directions aligned with the text subspace. The unique components $U_V$ and $U_t$ measure target-related information along modality-specific directions orthogonal to the other modality's estimated subspace. For synergy, we evaluate whether correctly paired image-text inputs yield more target-related information within $G_S$ than mismatched inputs. We generate $K$ mismatched conditions using independent arrangements of the images while keeping the texts and targets fixed, so that no original image-text pair remains. Let $H_{\mathrm{mis}}^{(k)}$ denote the decision representations from the $k$-th mismatched condition. Using the same geometry $G_S$ estimated from the matched reference batch for all conditions, we define
\begin{equation}
S
=
I_2(HG_S;Y)
-
\frac{1}{K}
\sum_{k=1}^{K}
I_2\left(H_{\mathrm{mis}}^{(k)}G_S;Y\right).
\label{eq:synergy}
\end{equation}
A positive value indicates greater target-related information under correct pairing, whereas a negative value indicates the reverse. Further image substitution test across 24 model $\times$ benchmark pairs to validate $S$ is done in Appendix~\ref{app:e-correspondence}.


\subsection{Analyzing Visual Information Use}
\label{sec:analysis}

\begin{figure}[ht!]
\centering
\includegraphics[scale =0.46]{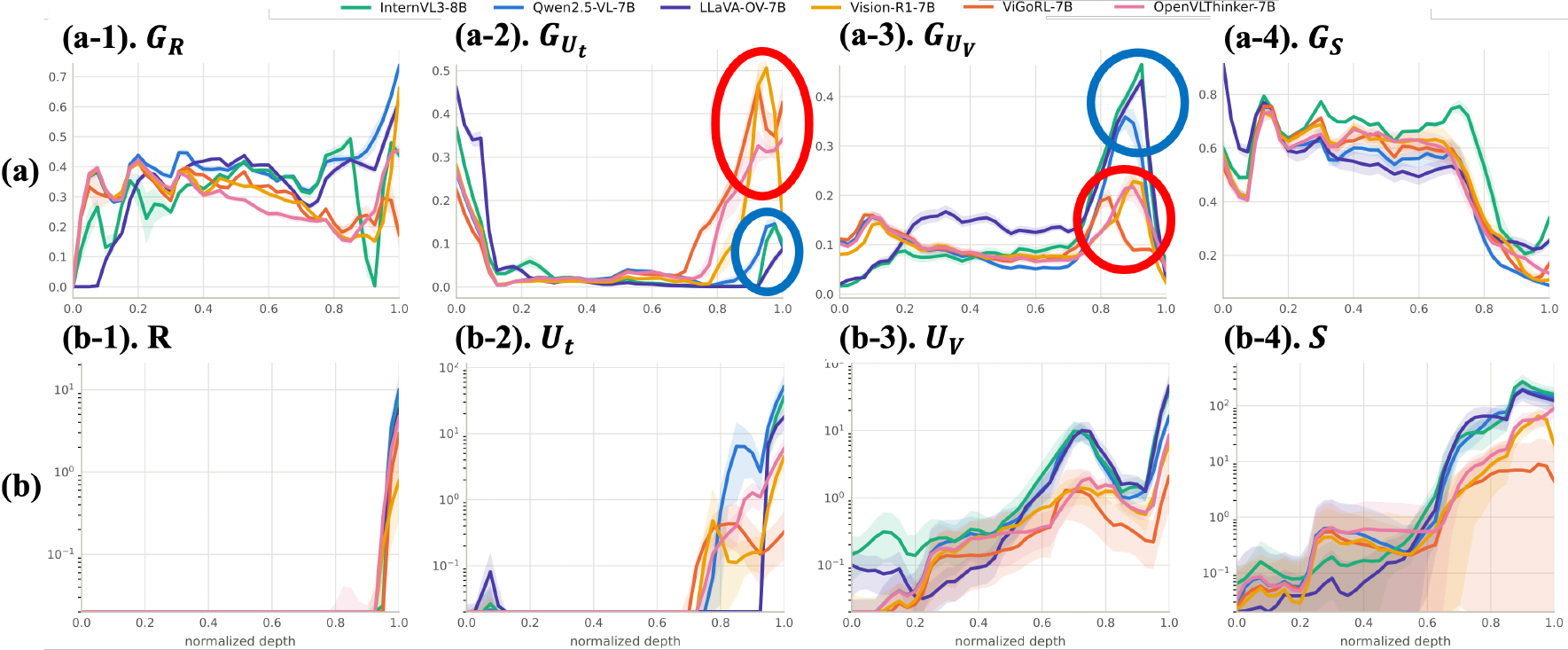}
\caption{(a) Layer-wise geometric measures for the shared, text-unique, vision-unique, and synergy-evaluation subspaces. Models highlighted in \textcolor{red}{red} and \textcolor{blue}{blue} exhibit contrasting text-unique and vision-unique profiles in later layers. (b) Models with more pronounced vision-unique components (\textcolor{blue}{blue}) also tend to show higher $U_V$ and $S$. The x-axis denotes normalized layer depth. Appendix~\ref{app:e-benchmarks} extends the information measurements in (b) across $14$ benchmarks.}
\label{fig:layerwise}
\vspace{-0.1cm}
\end{figure}

\paragraph{How PID information is organized across layers.}
\label{sec:exp-layers}
Figure~\ref{fig:layerwise} presents geometric profiles in the upper panels and their associated answer-information measurements in the lower panels across $6$ VLMs. We first examine the shared geometry in panel (a-1), which generally increases in the early layers. The modality-specific panels (a-2) and (a-3) reveal a contrasting pattern across models at later layers, particularly around normalized depths of $0.8$-$0.95$. The red circles highlight Vision-R1-7B~\citep{huang2026vision}, ViGoRL-7B~\citep{sarch2026grounded}, and OpenVLThinker-7B~\citep{deng2025openvlthinker}, which exhibit stronger text-unique and weaker vision-unique components than the models highlighted by the blue circles: InternVL3-8B~\citep{zhu2025internvl3}, Qwen2.5-VL-7B-Instruct~\citep{bai2025qwen25vltechnicalreport}, and LLaVA-OV-7B~\citep{li2024llava}. We compare this geometry with the actual answer-related information in Figure~\ref{fig:layerwise}(b). The models highlighted by the red circles in (a) also tend to exhibit lower vision-unique information $U_V$ (b-3) and $S$ (b-4) in later layers. This association between weaker vision-unique representations and lower $U_V$ motivates examining vision-unique directions ($G_{U_V}$) as candidates for amplification.

\paragraph{How do PID components vary with the information needed to answer a question?}
\label{sec:exp-evidence}

\begin{figure}[t]
\centering

\begin{minipage}[t]{0.48\linewidth}
\vspace{-0.2cm}
\centering
\small
\setlength{\tabcolsep}{5pt}
\medskip
\begin{tabular}{lcccc}
\toprule
& \multicolumn{4}{c}{Normalized by $\Irr(h;Y)$} \\
\cmidrule(lr){2-5}
Type & $R$ & $U_v$ & $U_t$ & $S$ \\
\midrule
$\mathbf{T}$
    & $0.18$ & $1.15$ & $0.92$ & $2.24$ \\
$\mathbf{T^-}$
    & $0.17$ & $1.17$ & $\mathbf{1.03}$ & $3.27$ \\
$\mathbf{V}$
    & $0.22$ & $\mathbf{1.34}$ & $0.90$ & $\mathbf{3.56}$ \\
$\mathbf{J}$
    & $0.19$ & $1.13$ & $1.00$ & $\mathbf{3.55}$ \\
\bottomrule
\end{tabular}
\captionof{table}{Normalized PID components across question types. \textbf{T}: questions answerable from text alone, $\mathbf{T^-}$: questions where text suggests a likely answer, \textbf{V}: questions requiring direct visual observation, \textbf{J}: questions requiring joint textual and visual information. Each component is divided by the question type's total decision-state information $\Irr(h;Y)$. Mutual information (Eq.~\ref{eq:renyi_mi}) is computed using 40 samples per each question type. This analysis covers $22$ VLMs and $11$ benchmarks (containing at least two question types with at least $40$ items each), yielding $253$ combinations.}
\label{tab:exp-evidence}
\end{minipage}
\hfill
\begin{minipage}[t]{0.49\linewidth}
\vspace{-0.2cm}
\centering
\includegraphics[
    width=0.67\linewidth
]{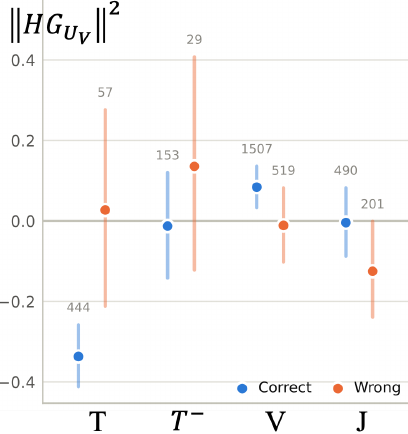}
\vspace{-0.2cm}
\captionof{figure}{Magnitude of the vision-unique component ($\Vert{}H G_{U_v}\Vert{}^2$), grouped by question type and prediction correctness for Qwen2.5-VL-7B~\citep{bai2025qwen25vltechnicalreport}. For visually demanding questions (\textbf{V}, \textbf{J}), correct predictions exhibit stronger vision-unique representations. Values are standardized within each benchmark. Statistical significance experiment is in Appendix~\ref{app:e-correctness}.}
\label{fig:item-guv}

\end{minipage}
\vspace{-0.4cm}
\end{figure}

To investigate this, we categorize the evaluation questions into four distinct types based on their evidence requirements by utilizing a LLM annotator (Claude Opus 5~\citep{anthropic2026claude}) to ensure objective categorization (detailed in Appendix~\ref{supp:question_annotation}). Text-sufficient (\textbf{T}) questions can be answered confidently using only the provided text and prior knowledge. Text-prior ($\mathbf{T^-}$) questions contain textual clues that strongly suggest an answer, though the image remains necessary for certainty. Visual-observation (\textbf{V}) questions require direct inspection of the image content, such as object identification or counting. Lastly, Joint-evidence (\textbf{J}) questions mandate synthesizing substantive textual premises with visual observations to deduce the answer. In Table~\ref{tab:exp-evidence}, comparing the $U_V$ column, \textbf{V} exhibits the highest value ($1.34$). Geometrically, $U_V$ captures purely visual features that are independent of the text. This shows that when a question requires visual grounding (i.e., question type \textbf{V}), the model correctly relies on this unique visual evidence. In the $S$ column, both \textbf{V} and \textbf{J} exceed \textbf{T} ($3.56$ and $3.55$, versus $2.24$). Geometrically, $S$ measures the joint information existing outside the isolated vision and text spaces. These high scores for both \textbf{V} and \textbf{J} demonstrate that visual grounding actively drives cross-modal interaction. Appendix~\ref{app:e-correlations} further shows that $S$ correlates with models' use of visual information. Overall, these patterns suggest that the PID components measured by \textsc{GeoPID} reflect whether answering a question requires textual information, visual information, or both, connecting representation geometry to the information the task requires.

\paragraph{How vision-unique components relate to prediction correctness.}
We further divide each category into correctly and incorrectly answered questions and compare the proportion of the decision representation along the vision-unique directions $G_{U_V}$. Figure~\ref{fig:item-guv} shows this comparison for Qwen2.5-VL-7B-Instruct. For \textbf{T} and $\mathbf{T^-}$, correctly answered questions exhibit weaker vision-unique components than incorrectly answered questions. The relationship reverses for \textbf{V} and \textbf{J}, where correct answers exhibit stronger vision-unique components. This contrast associates stronger vision-unique components with correct predictions specifically when visual information is required, suggesting that indiscriminate visual amplification could actually harm text-sufficient reasoning. We therefore selectively amplify visual representation components along $G_{U_V}$ during inference and evaluate whether this intervention improves predictions across multiple VLM models and multiple benchmarks.

\subsection{Targeted Intervention in Visual Representations}
\label{sec:geometry_intervention}

\paragraph{From analysis to intervention.}
Section~\ref{sec:analysis} identifies $G_{U_V}$ as a candidate for strengthening only the vision directions not explained by the estimated text subspace. For each model-benchmark pair, we use a calibration subset that is disjoint from the final evaluation set. The geometric basis \(G_{U_V}\) is estimated from the representations in the calibration subset without using ground-truth labels. Ground-truth labels are used only to select the intervention depth \(\ell^*\) and amplification strength \(\alpha\) on the calibration subset. Once selected, these settings are fixed for the final evaluation, and no label-dependent sample selection or gating is performed at inference time.

\paragraph{Vision-unique amplification.}
Let $V_i^{\ell^\star}$ denote the matrix of visual representations at the selected decoder layer $\ell^\star$. We update all vision tokens simultaneously:
\begin{equation}
    V_i^{\ell^\star}
    \leftarrow
    V_i^{\ell^\star}
    \left(I+\alpha G_{U_V}^{\ell^\star}
    \left(G_{U_V}^{\ell^\star}\right)^\top\right).
    \label{eq:vision_intervention}
\end{equation}
The matrix
$G_{U_V}^{\ell^\star}(G_{U_V}^{\ell^\star})^\top$
is the orthogonal projector onto the vision-unique geometry. Because of this, the intervention scales the component along the vision-unique subspace by $1+\alpha$ and preserves its orthogonal complement unchanged. Further analysis and efficient implementation is in Appendix~\ref{app:md-intervention}. The remaining decoder layers process the modified visual representations and can integrate the amplified content into the prediction.

\section{Experiments}
\label{sec:exp}
We evaluate \textsc{GeoPID} across $22$ VLMs and $14$ benchmarks, yielding $308$ model $\times$ benchmark combinations. Experimental Details are in Appendix~\ref{supp:experimental_setup}. More results are in Appendix~\ref{supp:results}.

\begin{table}[t]\centering\small\setlength{\tabcolsep}{3pt}
\caption{Accuracy gain ($\%$) of the $G_{U_V}$ intervention at each cell's selected $(\ell,\alpha)$, 308 models $\times$ benchmarks. Benchmark abbreviation follows Appendix~\ref{app:benchmark-suite}.}
\label{tab:percell-qv-filtered}
\resizebox{\textwidth}{!}{%
\begin{tabular}{lrrrrrrrrrrrrrr}
\toprule
Model & POPE & MME & CVB & Hall & Reef & NatB & RWQA & MMB & SEED & AI2D & AOK & SQA & Rap4o & GenAI \\
\midrule
Qwen2.5-VL-7B~\citep{bai2025qwen25vltechnicalreport} & +1.0 & +0.0 & +2.9 & +3.6 & +0.5 & +2.7 & +0.7 & +0.2 & +1.0 & +1.0 & +0.2 & +2.5 & +0.8 & +4.7 \\
Qwen3-VL-8B~\citep{bai2025qwen3} & +0.5 & +3.1 & +0.0 & +0.6 & +0.2 & +0.0 & +0.0 & +0.0 & -0.5 & +1.0 & +0.2 & +0.2 & +5.9 & +4.3 \\
InternVL3-8B~\citep{zhu2025internvl3} & +0.2 & -0.8 & +1.6 & +4.2 & +0.0 & +1.3 & +1.3 & -0.5 & +0.7 & +0.5 & -0.2 & -1.0 & +2.3 & +11.0 \\
InternVL3-2B~\citep{zhu2025internvl3} & +0.2 & +1.8 & +2.3 & +0.6 & +0.2 & +3.8 & +4.7 & +1.1 & +0.7 & -0.8 & +0.2 & -0.5 & +1.1 & -5.7 \\
LLaVA-OV-7B~\citep{li2024llava} & +1.4 & +0.8 & -0.8 & +3.0 & +1.2 & +0.5 & -1.3 & +0.0 & +0.7 & -0.3 & +0.9 & +0.7 & +27.1 & +2.7 \\
Idefics2-8B~\citep{laurenccon2024matters} & +0.0 & +0.5 & +1.8 & -1.2 & +2.6 & +1.1 & +1.3 & +0.2 & +0.0 & +1.6 & +1.1 & -0.5 & -1.1 & -8.7 \\
SmolVLM-2B~\citep{marafioti2025smolvlm} & +0.5 & -1.3 & +3.1 & +3.6 & +0.7 & +1.9 & +0.7 & +0.5 & -0.7 & +0.3 & -0.5 & +2.9 & +4.0 & +9.3 \\
MMR1-Math-7B~\citep{leng2025mmr1} & +2.4 & +0.0 & +0.0 & -0.6 & +0.0 & +1.3 & +0.0 & -0.2 & +0.0 & +1.8 & -0.5 & +2.5 & +0.8 & +7.3 \\
VL-Rethinker-7B~\citep{wang2026vl} & +2.1 & +2.3 & +1.0 & +1.2 & -0.2 & +0.3 & +0.0 & +0.2 & -0.2 & +1.0 & +0.7 & +0.2 & +8.5 & -0.3 \\
Orsta-7B~\citep{ma2025one} & +0.2 & +1.0 & +0.8 & +2.4 & +0.2 & +1.1 & +1.3 & +0.5 & +1.2 & +0.3 & +0.0 & +1.5 & +0.3 & +6.0 \\
ThinkLite-VL-7B~\citep{wang2026sota} & +0.5 & +2.6 & +2.6 & +1.2 & -0.2 & +1.1 & -0.7 & +0.0 & -0.2 & +1.3 & +0.0 & +1.0 & +2.0 & +4.7 \\
VLAA-Thinker-Q2.5VL-7B~\citep{chen2025sft} & +1.2 & +1.0 & -0.5 & +0.0 & +1.0 & +2.4 & -0.7 & +0.0 & +0.7 & +1.6 & +0.2 & +2.7 & +5.1 & +7.0 \\
VLAA-Thinker-Q2.5VL-3B~\citep{chen2025sft} & +1.9 & +0.0 & +0.8 & +3.6 & +0.2 & +2.7 & +1.3 & +0.0 & +1.2 & +1.0 & +0.2 & +0.7 & +14.1 & +7.0 \\
VLAA-Thinker-Q2VL-7B~\citep{chen2025sft} & +0.5 & -0.3 & +2.1 & +4.2 & +0.2 & +1.1 & +0.7 & +0.7 & +2.2 & -0.3 & +0.2 & +1.0 & +4.5 & +2.3 \\
VLAA-Thinker-Q2VL-2B~\citep{chen2025sft} & +0.2 & +0.3 & +1.8 & +0.0 & -0.2 & -0.5 & -1.3 & +0.2 & +0.5 & +0.8 & +0.5 & -1.5 & +11.0 & -0.3 \\
VLM-R1-OVD-3B~\citep{shen2025vlm} & +0.5 & +0.8 & +1.0 & +1.8 & +0.0 & +5.4 & +2.0 & +0.0 & +0.7 & +1.3 & +0.0 & +1.0 & +17.8 & +12.0 \\
VLM-R1-Math-3B~\citep{shen2025vlm} & +1.7 & +0.8 & +1.3 & +3.0 & +1.0 & +5.1 & +2.7 & +0.0 & +0.7 & +1.0 & -0.2 & +0.7 & +5.9 & +7.7 \\
OpenVLThinker-7B~\citep{deng2025openvlthinker} & +1.0 & +3.6 & +1.3 & -0.6 & +0.5 & +0.5 & -0.7 & +0.5 & +0.7 & -0.8 & +0.9 & +1.2 & -4.0 & +0.3 \\
Vision-R1-7B~\citep{huang2026vision} & +0.5 & +1.0 & -0.5 & +3.0 & +4.1 & -0.3 & +1.3 & -0.2 & -0.7 & +0.8 & +1.1 & +1.0 & +52.8 & +13.7 \\
ViGoRL-7B~\citep{sarch2026grounded} & +32.1 & +42.7 & +39.7 & +37.3 & +60.6 & +43.4 & +62.4 & +59.8 & +29.5 & +59.6 & +58.2 & +46.3 & +69.8 & +26.7 \\
ViGoRL-3B~\citep{sarch2026grounded} & +18.6 & +30.8 & +31.9 & +30.7 & +3.1 & +2.4 & +45.6 & +5.7 & +49.9 & +18.1 & +12.9 & +6.1 & +75.7 & +15.3 \\
ViGoRL-MCTS-3B~\citep{sarch2026grounded} & +0.7 & +26.4 & +6.0 & +24.7 & +0.7 & +6.4 & +38.3 & +1.4 & +50.1 & +16.6 & +3.2 & +0.0 & +91.2 & +31.7 \\
\bottomrule
\end{tabular}
}
\vspace{-0.4cm}
\end{table}
\begin{table}[t]\centering\small
\begin{minipage}[t]{0.38\textwidth}\centering
\caption{Accuracy before and after the $G_{U_V}$ intervention (Table~\ref{tab:percell-qv-filtered}) by four question type (\textbf{T}, $\mathbf{T^-}$, \textbf{V, J}) defined in Appendix~\ref{supp:question_annotation} (gain in $\%$).}
\label{tab:census-types-filtered}
\setlength{\tabcolsep}{5pt}
\begin{tabular}{lccc}
\toprule
Type & Before & After & Gain \\
\midrule
\textbf{T}   & $80.1$ & $83.6$ & $+3.5$ \\
$\mathbf{T^-}$   & $69.4$ & $73.1$ & $+3.7$ \\
\textbf{V}   & $69.4$ & $73.6$ & $+4.2$ \\
\textbf{J}   & $64.9$ & $74.5$ & $\mathbf{+9.6}$ \\
\midrule
All & $70.3$ & $75.4$ & $+5.1$ \\
\bottomrule
\end{tabular}
\end{minipage}\hfill
\begin{minipage}[t]{0.58\textwidth}\centering
\caption{Comparison of \textsc{GeoPID} against baseline methods. \emph{Acc.} represents accuracy (\%), \emph{Gain} indicates absolute improvement ($\%$), \emph{Corr.} denotes $\%$ of incorrect answers fixed, while \emph{Dam.} denotes correct answers broken.}
\label{tab:baseline_comparison}
\setlength{\tabcolsep}{4pt}
\begin{tabular}{lcccc}
\toprule
Method & Acc. & Gain & Corr. & Dam. \\
\midrule
M3ID~\citep{favero2024multi} & 76.5 & +2.54 & 21.2 & 5.6 \\
VCD~\citep{leng2024mitigating} & 75.6 & +1.65 & 18.5 & 6.0 \\
VTI~\citep{liu2024reducing} & 72.9 & -1.04 & 9.2 & \textbf{4.6} \\
DoLa~\citep{chuang2024dola} & 71.0 & -2.91 & 18.9 & 11.8 \\
LoRA-PID~\citep{fang2026towards} & 74.87 & +2.50 & -- & -- \\
\textbf{\textsc{GeoPID}} & \textbf{78.9} & \textbf{+4.92} & \textbf{21.6} & 4.8 \\
\bottomrule
\end{tabular}
\end{minipage}
\vspace{-0.4cm}
\end{table}
\subsection{Can \textsc{GeoPID} Improve Predictions?}
\label{sec:exp-intervention}
We apply the intervention Eq.~\ref{eq:vision_intervention} to amplify representation components along the vision-unique subspace and evaluate whether this improves predictions. For each of the 308 model $\times$ dataset pairs, we reserve 100 questions as a calibration subset, disjoint from the final evaluation set. We use the ground-truth labels of these 100 questions to select the $\ell^*/L\in [0.3,0.8]$ and $\alpha\in {1,2,4,6,8,12}$ by maximizing calibration accuracy. The selected ($\ell^*,\alpha$) is then fixed and applied to the full evaluation set without further tuning or label-dependent selection. Table~\ref{tab:percell-qv-filtered} reports the resulting accuracy gains, yielding an average relative improvement of 7.63\%. This shows that the subspace identified by the analysis can serve as an effective intervention target without updating model parameters. Table~\ref{tab:census-types-filtered} reports accuracy before and after intervention for the four question types. The largest improvement occurs for questions requiring textual and visual information jointly (\textbf{J}). This is consistent with Figure~\ref{fig:item-guv}: strengthening vision-unique components can help when the model must use visual observations, or must combine visual with textual information. Moreover, positive gains for \textbf{T} and $\mathbf{T^-}$ also show that gains are not confined to the visually demanding questions alone.

The gains in Table~\ref{tab:percell-qv-filtered} vary across models and benchmarks. This variability motivates us to investigate whether a model's baseline geometric structure can guide its optimal intervention settings.

\begin{figure}[ht!]
\centering
\includegraphics[scale =0.5]{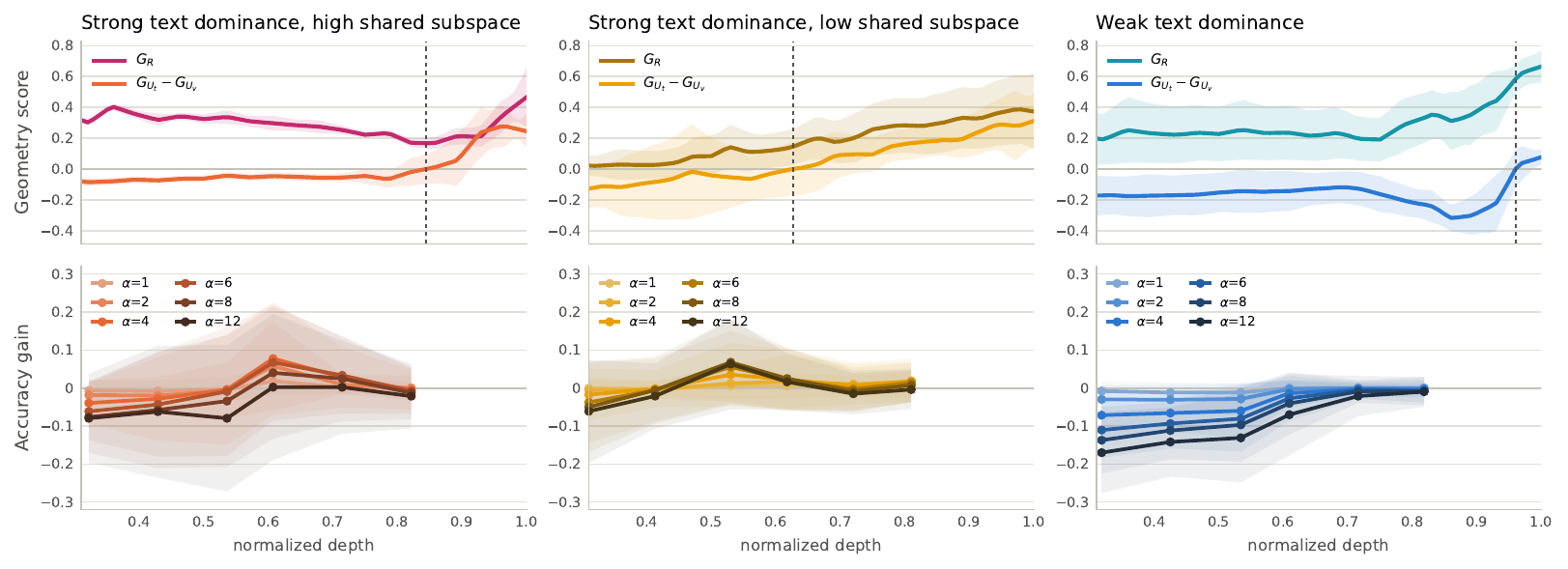}
\caption{Geometry-based intervention effects across three regimes: strong late-layer text-unique $G_{U_t}$ dominance with high shared geometry $G_R$ (left), strong text-unique dominance with low shared geometry (middle), and weak text-unique dominance (right). Top: layer-wise profiles of shared geometry $G_R$ and the difference $G_{U_t}-G_{U_v}$. Bottom: mean accuracy gains from the intervention in Eq.~\ref{eq:vision_intervention} at different $\ell^*$ and $\alpha$. The strong text dominance regimes favor early interventions with high amplification strengths, whereas the weak text dominance regime shows degradation under strong early interventions.}
\label{fig:rule}
\vspace{-0.5cm}
\end{figure}
\subsection{Can Geometry Guide Intervention Settings?}
\label{sec:exp-rule}

\begin{table}[t]\centering\small\setlength{\tabcolsep}{3pt}
\caption{Accuracy gain ($\%$) of the $G_{U_v}$ intervention at the $(\ell^*,\alpha)$ chosen by a fixed geometry-based rule described in Section~\ref{sec:exp-rule}, across 22 models $\times$ 14 datasets. The thresholds were selected using the same 100-question calibration subset and then fixed. Benchmark abbreviation is in Appendix~\ref{app:benchmark-suite}.}
\label{tab:percell-geotree_d4-filtered}
\resizebox{\textwidth}{!}{%
\begin{tabular}{lrrrrrrrrrrrrrrr}
\toprule
Model & POPE & MME & CVB & Hall & Reef & NatB & RWQA & MMB & SEED & AI2D & AOK & SQA & Rap4o & GenAI & Mean \\
\midrule

Qwen2.5-VL-7B~\citep{bai2025qwen25vltechnicalreport} & -0.2 & +0.0 & -0.5 & +0.6 & +0.2 & +2.7 & -0.7 & +0.5 & +0.7 & +1.3 & +0.0 & +2.5 & +0.8 & +4.7 & +1.05 \\
Qwen3-VL-8B~\citep{bai2025qwen3} & +0.0 & +0.0 & +0.0 & +0.6 & +0.0 & +0.0 & +0.7 & +0.0 & -1.7 & +0.8 & +0.0 & +0.2 & +5.1 & +0.7 & +0.49 \\
InternVL3-8B~\citep{zhu2025internvl3} & +0.5 & +0.3 & +0.5 & +0.6 & +0.0 & +0.0 & +1.3 & -0.2 & +0.5 & +0.3 & +0.0 & +0.0 & -1.1 & +0.0 & +0.15 \\
InternVL3-2B~\citep{zhu2025internvl3} & +0.0 & -0.3 & +0.3 & +0.6 & +0.2 & +1.6 & +1.3 & +0.0 & +0.5 & -0.8 & -0.2 & +0.0 & +1.1 & +2.0 & +0.59 \\
LLaVA-OV-7B~\citep{li2024llava} & +0.7 & +0.5 & -0.8 & +1.2 & +0.7 & +0.5 & -2.7 & +0.5 & +1.0 & -0.3 & +0.5 & +0.7 & +27.4 & -6.7 & +1.46 \\
Idefics2-8B~\citep{laurenccon2024matters} & +0.5 & +0.0 & +1.6 & +0.0 & -0.2 & +0.8 & +0.7 & -0.5 & +0.7 & +0.3 & +0.5 & -0.5 & -1.7 & -1.0 & +0.17 \\
SmolVLM-2B~\citep{marafioti2025smolvlm} & +0.7 & -0.8 & -0.3 & +4.2 & -0.2 & -2.1 & +0.0 & +0.0 & -0.7 & +1.6 & -0.5 & -0.5 & +1.1 & +2.0 & +0.33 \\
MMR1-Math-7B~\citep{leng2025mmr1} & +0.0 & -0.3 & +0.0 & +0.6 & +0.0 & +1.3 & +0.0 & +0.5 & +0.7 & +1.3 & +0.0 & +0.0 & +1.1 & +7.3 & +0.97 \\
VL-Rethinker-7B~\citep{wang2026vl} & +0.0 & -0.3 & +1.0 & -0.6 & -0.2 & -0.3 & -1.3 & +0.2 & +0.2 & +0.0 & +0.0 & +0.2 & +3.7 & +0.0 & +0.21 \\
Orsta-7B~\citep{ma2025one} & +0.0 & +0.0 & +0.8 & +0.0 & -0.2 & -0.3 & +0.7 & +0.5 & +0.2 & +0.3 & -0.7 & +1.5 & +2.5 & +6.0 & +1.05 \\
ThinkLite-VL-7B~\citep{wang2026sota} & +0.0 & +0.3 & +0.3 & +0.0 & -0.2 & +1.3 & +0.0 & +0.0 & +0.5 & +1.3 & +0.5 & +0.0 & +2.0 & +4.7 & +0.81 \\
VLAA-Thinker-Q2.5VL-7B~\citep{chen2025sft} & +0.0 & +0.8 & +0.0 & +0.0 & +0.0 & +2.4 & -0.7 & +0.7 & +0.5 & +0.3 & -0.2 & +2.7 & +1.4 & +3.3 & +0.95 \\
VLAA-Thinker-Q2.5VL-3B~\citep{chen2025sft} & +0.5 & +0.5 & +1.6 & -1.2 & -0.2 & +0.3 & +0.0 & +0.0 & +1.0 & +0.0 & -0.2 & +0.2 & +14.1 & +0.3 & +1.01 \\
VLAA-Thinker-Q2VL-7B~\citep{chen2025sft} & +0.2 & +0.3 & +0.0 & -0.6 & +0.2 & +2.1 & +0.7 & +0.5 & +0.5 & -0.3 & +0.2 & +1.0 & +4.5 & +3.7 & +0.65 \\
VLAA-Thinker-Q2VL-2B~\citep{chen2025sft} & +0.5 & +0.3 & +2.1 & -1.8 & +0.5 & -0.8 & +0.0 & +0.0 & +0.5 & +0.8 & +0.2 & -0.2 & +7.9 & -5.0 & +0.49 \\
VLM-R1-OVD-3B~\citep{shen2025vlm} & -0.2 & +0.8 & -0.3 & -0.6 & -0.2 & -0.8 & -0.7 & +0.0 & +0.0 & +1.3 & +0.2 & +0.2 & +17.8 & -1.0 & +1.08 \\
VLM-R1-Math-3B~\citep{shen2025vlm} & +0.0 & +0.3 & +0.3 & +1.8 & +0.2 & +0.5 & +0.0 & +0.0 & +0.0 & -0.3 & -0.9 & -0.2 & +5.9 & +0.7 & +0.61 \\
OpenVLThinker-7B~\citep{deng2025openvlthinker} & +0.0 & +0.0 & -1.0 & -0.6 & +0.5 & +0.5 & -0.7 & -0.2 & -0.5 & +0.0 & -1.4 & -1.0 & -0.6 & -1.3 & +0.09 \\
Vision-R1-7B~\citep{huang2026vision} & +0.2 & -0.3 & +1.0 & +1.8 & +4.1 & -0.3 & +2.7 & -1.6 & -0.7 & -0.8 & +6.9 & -2.2 & +51.7 & -0.3 & +3.40 \\
ViGoRL-7B~\citep{sarch2026grounded} & +39.3 & +42.7 & +39.7 & +37.3 & +60.6 & +43.4 & +62.4 & +48.2 & +60.7 & +59.6 & +58.2 & +41.2 & +55.1 & +48.3 & +45.47 \\
ViGoRL-3B~\citep{sarch2026grounded} & +26.9 & +30.8 & +16.1 & +30.7 & +13.2 & +6.4 & +46.3 & +5.7 & +49.9 & +23.3 & +12.6 & +6.1 & +75.7 & +30.7 & +23.59 \\
ViGoRL-MCTS-3B~\citep{sarch2026grounded} & +26.4 & +26.4 & +11.4 & +24.7 & +19.5 & +11.8 & +38.3 & +0.5 & +50.1 & +16.6 & +15.2 & +0.5 & +91.2 & +31.7 & +21.86 \\
\bottomrule
\end{tabular}
}
\vspace{-0.5cm}
\end{table}
The effectiveness of vision-unique amplification correlates strongly with the base models' geometric profiles. Appendix~\ref{supp:exp-rule-details} shows that the layer-wise trajectory of $G_R$, $G_{U_t}$, $G_{U_v}$, and $S$ shows a strong association with intervention gain with up to pearson correlation $r=0.80$. Motivated by this predictive capability, we derive a fixed intervention rule based on the three geometric patterns identified in Figure~\ref{fig:rule} . For representations exhibiting strong late-layer text dominance $(G_{U_t} - G_{U_v} \ge 0.1)$ alongside a broad early shared geometry $(G_R \ge 0.296)$, a moderate visual amplification $(l^*/L=0.6, \alpha=4)$ is optimal. When this early shared space is weak, earlier and stronger amplification $(0.5, 8)$ is required to integrate the modalities. Conversely, representations with weak text dominance naturally maintain geometric balance and necessitate only minimal, late-stage amplification $(0.7, 1)$. Applying this deterministic rule across all 308 model-benchmark pairs (Table~\ref{tab:percell-geotree_d4-filtered}) yields a relative accuracy gain of 4.62\%. Furthermore, compared to established training-free baselines~\citep{favero2024multi, leng2024mitigating, liu2024reducing, chuang2024dola} and a training-based PID baseline~\citep{fang2026towards} in Table~\ref{tab:baseline_comparison}, \textsc{GeoPID} achieves the highest overall accuracy and absolute gain while maintaining a minimal damage rate. Unlike global or heuristic shifts that often disrupt pre-existing knowledge, selectively targeting the vision-unique subspace effectively balances error correction with capability preservation.


\section{Conclusion}
We present a framework to geometrically analyze and steer how VLMs distribute and utilize multimodal information via \textsc{GeoPID}. Specifically, adjusting the depth and strength of the intervention based on early shared geometry and text dominance leads to consistent gains across diverse benchmarks. Ultimately, our intervention shows that appropriately configured subspace manipulation can reliably enhance model performance. This framework provides a useful tool for analyzing multimodal representation spaces and addressing reasoning bottlenecks without costly retraining.
\subsection*{AI use statement}

In this work, we used generative AI tools for supporting qualitative data analysis, specifically for categorizing the evaluation questions based on their evidence requirements as detailed in Appendix~\ref{supp:question_annotation}. We have not used generative AI tools for generating synthetic data sets, formulating theoretical models, or writing mathematical proofs, and the rest of the required disclosure tasks are not applicable to this work. Additionally, we used generative AI tools to polish writing to improve readability and format LaTeX tables. We have reviewed all AI-assisted work. Specifically, the question categorizations generated by Claude Opus 5~\citep{anthropic2026claude} were subjected to manual review, and all AI-formatted LaTeX tables and text revisions were verified for accuracy by the authors. We take responsibility for the final content of this work, including text, claims, or artifacts produced with the aid of generative AI.

\subsection*{Ethics statement}

In conducting this research, we utilized publicly available vision-language benchmarks and evaluated explicitly grounded models. We ensured strict adherence to data compliance protocols, particularly regarding the handling of potential personally identifiable information that may be present in real-world image datasets. All model and dataset usage was reviewed for proper licensing and ethical compliance. We anticipate no direct societal harm from the \textsc{GeoPID} framework, as it serves as a training-free inference intervention to improve visual grounding and mitigate hallucination, rather than a generative source of new multimodal content.

\subsection*{Reproducibility statement}

It is important that the work published in ICLR is reproducible. While the source code cannot be publicly released at this time, we ensure full reproducibility by explicitly detailing all necessary experimental configurations, implementations, and mathematical formulations within the manuscript. Specifically, the exact construction of the geometric subspaces and the complementary projection are fully specified via equations in Section~\ref{sec:method} and Appendix~\ref{sec:detailed-method}. All hyperparameter settings, including the rank ($r$) for batch-wise PCA, intervention depth ($l^*$), and amplification strength ($\alpha$), are comprehensively documented in Section~\ref{sec:exp} and Appendix~\ref{supp:experimental_setup}. Furthermore, all evaluations were conducted using strictly publicly available vision-language benchmarks and open-weight models, ensuring that independent researchers can fully reconstruct and verify the \textsc{GeoPID} framework based solely on the provided theoretical formulations and procedural details.
\bibliography{iclr2027_conference}
\bibliographystyle{iclr2027_conference}

\appendix
\clearpage
\setcounter{page}{1}
\section{Supplementary}
\noindent The appendix includes the following sections:
\begin{itemize}
    \item \S~\ref{lab:related-works} Related Works
    \item \S~\ref{supp:experimental_setup} Experimental Setup
    \item \S~\ref{supp:question_annotation} Question Category Annotation
    \item \S~\ref{sec:detailed-method} More Analysis
    \item \S~\ref{supp:results} More Results
    \item \S~\ref{supp:ablations} More Ablation Studies
    \item \S~\ref{supp:robustness} Robustness
\end{itemize}

\section{Related Works}
\label{lab:related-works}
\subsection{Analyzing multimodal interaction with PID}
Partial Information Decomposition (PID)~\citep{williams2010nonnegative} splits the information that two modalities provide about a task into redundant, unique, and synergistic parts. A line of work develops estimators for these quantities, from unique information definitions~\citep{bertschinger2014quantifying}
and dedicated PID solvers~\citep{makkeh2018broja} to neural mutual-information estimators~\citep{belghazi2018mutual, oord2018representation}. Building on these, 
~\citet{liang2023multimodal, liang2023quantifying} proposed scalable estimators that quantify multi-modal datasets and models from this perspective. The framework was then moved to use MLLMs~\citep{fang2026towards, wu2026vision}. These works either decompose the modality contributions at the decision level~\citep{fang2026towards}, extend the PID decomposition layer by layer~\citep{wu2026vision}, or profile the information spectrum across 26 large VLMs~\citep{xiu2026comprehensive}. Beyond diagnosis, several works attempt an intervention, either by teacher-student distillation~\citep{liangvision}, or by fine-tuning with PID-guided reweighting~\citep{fang2026towards}. However, these approaches share two limitations. First, they need a separate estimator to obtain the PID components, whose output is only approximate for continuous, high-dimensional representations. Second, any improvements they report are obtained through additional training. Unlike these, we read the information directly from the representation matrices via their rank, without assuming any distribution. We further turn the resulting diagnosis into a performance gain through a training-free intervention.

\subsection{Representation Geometry and Information Measurement}
The rank of a representation, i.e., how many independent directions it spans, has been used as a measure of a matrix's information content and representational diversity~\citep{7098875, renyi1961measures, kokilepersaud2025hex, kokilepersaud2025adadim}. Recent multi-modal fusion work uses it to diagnose modality contributions and modality collapse~\citep{kim2026countering, chaudhuri2025closer, yarici2026mer}. These works treat every direction as equally useful and simply raise the total rank of the representation. However, the overall rank does not distinguish directions shared across modalities from those specific to each modality. We identify shared and modality-specific subspaces through the geometric relationships between visual and textual representations, and measure the answer-related information associated with these subspaces. This analysis identifies vision-unique directions as candidates for selective amplification.

\subsection{Training-free Interventions for Visual Information Use}
A large body of work reduces VLM hallucination in a training-free manner. These methods either contrast the logits of a removed visual input to amplify the vision contribution (contrastive decoding)~\citep{leng2024mitigating, chen2025curing}, shift activations toward visual-perception directions~\citep{liu2024reducing, yin2026dynamic}, intervene at the token level~\citep{li2026geass, wang2025mint}, or adjust the information flow~\citep{liu2026aligning, zhao2025mitigating, mao2026through, wang2025vegas, kim2026information}. However, these methods do not explicitly use PID to identify the intervention target. In contrast, our approach uses geometric PID analysis to identify a vision-unique subspace and selectively amplifies visual representation components along vision-unique subspace during inference.

\section{Experimental Setup}
\label{supp:experimental_setup}
\subsection{Models}
\label{app:model-suite}
The model suite includes general-purpose instruction-tuned VLMs, reasoning-oriented models, and models developed for explicit visual grounding. Their nominal sizes range from approximately $2$B to $8$B parameters. The evaluated model weights remain fixed throughout these experiments.

\begingroup
\small
\setlength{\tabcolsep}{3pt}
\renewcommand{\arraystretch}{1.15}
\begin{longtable}{@{}p{0.35\linewidth}p{0.04\linewidth}p{0.18\linewidth}p{0.37\linewidth}@{}}
\caption{The $22$ evaluated VLMs. Family denotes the multimodal model family or parent VLM.}\label{tab:models}\\
\toprule
Model & Size & Family / parent & Model type and emphasis \\
\midrule
\endfirsthead
\multicolumn{4}{l}{\tablename~\thetable\ (continued)}\\
\toprule
Model & Size & Family / parent & Model type and emphasis \\
\midrule
\endhead
\midrule
\multicolumn{4}{r}{Continued on next page}\\
\endfoot
\bottomrule
\endlastfoot
\multicolumn{4}{l}{\textbf{General-purpose instruction-tuned VLMs}}\\
Qwen2.5-VL-7B~\citep{bai2025qwen25vltechnicalreport} & 7B & Qwen2.5-VL & Image and document understanding with dynamic-resolution visual processing. \\
Qwen3-VL-8B~\citep{bai2025qwen3} & 8B & Qwen3-VL & General multimodal instruction following; multi-level visual feature integration. \\
InternVL3-8B~\citep{zhu2025internvl3} & 8B & InternVL3 & Native multimodal pre-training with broad visual and linguistic capabilities. \\
InternVL3-2B~\citep{zhu2025internvl3} & 2B & InternVL3 & Smaller-scale member of the same model family. \\
LLaVA-OV-7B~\citep{li2024llava} & 7B & LLaVA-OneVision & Unified single-image, multi-image, and video understanding. \\
Idefics2-8B~\citep{laurenccon2024matters} & 8B & Idefics2 & Interleaved image--text understanding with pooled visual representations. \\
SmolVLM-2B~\citep{marafioti2025smolvlm} & $\sim$2B & SmolVLM & Compact instruction-tuned VLM with compressed visual representations. \\
\midrule
\multicolumn{4}{l}{\textbf{Reasoning- and task-oriented post-trained VLMs}}\\
MMR1-Math-7B~\citep{leng2025mmr1} & 7B & Qwen2.5-VL & Mathematical visual reasoning through GRPO-based post-training. \\
VL-Rethinker-7B~\citep{wang2026vl} & 7B & Qwen2.5-VL & RL designed to encourage reflection and verification during reasoning. \\
OpenVLThinker-7B~\citep{deng2025openvlthinker} & 7B & Qwen2.5-VL & Iterative SFT and RL for multimodal chain-of-thought reasoning. \\
Orsta-7B~\citep{ma2025one} & 7B & Qwen2.5-VL & Unified RL across reasoning and visual perception tasks. \\
ThinkLite-VL-7B~\citep{wang2026sota} & 7B & Qwen2.5-VL & Data-efficient reinforcement fine-tuning with MCTS-guided sample selection. \\
VLAA-Thinker-Q2.5VL-7B~\citep{chen2025sft} & 7B & Qwen2.5-VL & Reasoning-oriented post-training with perceptual and cognitive reward signals. \\
VLAA-Thinker-Q2.5VL-3B~\citep{chen2025sft} & 3B & Qwen2.5-VL & Smaller Qwen2.5-VL variant in the VLAA-Thinker suite. \\
VLAA-Thinker-Q2VL-7B~\citep{chen2025sft} & 7B & Qwen2-VL & Qwen2-VL variant in the VLAA-Thinker suite. \\
VLAA-Thinker-Q2VL-2B~\citep{chen2025sft} & 2B & Qwen2-VL & Smaller Qwen2-VL variant in the VLAA-Thinker suite. \\
VLM-R1-OVD-3B~\citep{shen2025vlm} & 3B & Qwen2.5-VL & RL specialization for open-vocabulary object detection. \\
VLM-R1-Math-3B~\citep{shen2025vlm} & 3B & Qwen2.5-VL & RL specialization for multimodal mathematical reasoning. \\
Vision-R1-7B~\citep{huang2026vision} & 7B & Qwen2.5-VL & Multimodal reasoning using a reasoning-data cold start followed by RL. \\
\midrule
\multicolumn{4}{l}{\textbf{Explicitly grounded reasoning models}}\\
ViGoRL-7B~\citep{sarch2026grounded} & 7B & Qwen2.5-VL & Visually grounded reasoning with RL. \\
ViGoRL-3B~\citep{sarch2026grounded} & 3B & Qwen2.5-VL & Smaller-scale visually grounded RL model. \\
ViGoRL-MCTS-SFT-3B~\citep{sarch2026grounded} & 3B & Qwen2.5-VL & SFT on MCTS-generated, spatially grounded reasoning traces. \\
\end{longtable}
\endgroup

\subsection{Benchmarks}
\label{app:benchmark-suite}
The benchmark suite covers broad multimodal understanding (object recognition, spatial understanding), visual perception and hallucination, scientific, knowledge-based, and medical reasoning, and generated-image comparison. This diversity allows us to examine visual information use under different model architectures, post-training objectives, and task requirements. Our evaluation uses a common answer-choice scoring protocol, evaluated in a multiple-choice format including binary questions and pairwise image comparisons. For multiple-choice evaluation, we compare the model’s output logits for the  answer choices and select the choice with the highest logit. We use identical prompts before and after intervention.

\begingroup
\small
\setlength{\tabcolsep}{3pt}
\renewcommand{\arraystretch}{1.15}
\begin{longtable}{@{}p{0.22\linewidth}p{0.08\linewidth}p{0.42\linewidth}p{0.20\linewidth}@{}}
\caption{The $14$ evaluated benchmarks. The final column describes the answer format used in this study. Dataset codes follow the complete appendix tables.}\label{tab:datasets}\\
\toprule
Benchmark & Code & Task and visual content & Evaluation format \\
\midrule
\endfirsthead
\multicolumn{4}{l}{\tablename~\thetable\ (continued)}\\
\toprule
Benchmark & Code & Task and visual content & Evaluation format \\
\midrule
\endhead
\midrule
\multicolumn{4}{r}{Continued on next page}\\
\endfoot
\bottomrule
\endlastfoot
\multicolumn{4}{l}{\textbf{Broad multimodal understanding}}\\
MME~\citep{fu2026mme} & MME & Perception and cognition, including recognition and reasoning. & Binary answer selection \\
MMBench~\citep{liu2024mmbench} & MMB & Multiple visual perception and reasoning abilities. & Multiple-choice selection \\
SEED-Bench~\citep{li2023seed} & SEE & Visual comprehension across several evaluation dimensions. & Multiple-choice selection \\
\midrule
\multicolumn{4}{l}{\textbf{Visual perception and hallucination}}\\
POPE~\citep{li2023evaluating} & POP & Presence or absence of objects in natural images. & Binary answer selection \\
HallusionBench~\citep{guan2024hallusionbench} & HAL & Image-context reasoning, language hallucination, and visual illusion. & Binary answer selection \\
CV-Bench~\citep{tong2024cambrian} & CVB & Counting and spatial relations in 2D; depth and relative distance in 3D. & Multiple-choice selection \\
NaturalBench~\citep{li2024naturalbench} & NAT & Natural image--question pairs designed to expose visual reasoning errors. & Answer-choice selection \\
RealWorldQA~\citep{realworldqa2024} & RWQ & Understanding real scenes and spatial relationships. & Multiple-choice subset \\
Reefknot~\citep{zheng2025reefknot} & REE & Perceptive and cognitive relations, including relation hallucinations. & Answer-choice subset \\
\midrule
\multicolumn{4}{l}{\textbf{Scientific, knowledge-based, and medical reasoning}}\\
AI2D~\citep{kembhavi2016diagram} & AI2 & Scientific diagrams, labels, components, and their relationships. & Multiple-choice selection \\
ScienceQA~\citep{lu2022learn} & SCI & Science questions with visual and textual context. & Image-containing MC subset \\
A-OKVQA~\citep{schwenk2022okvqa} & AOK & Questions combining natural images with commonsense or world knowledge. & Multiple-choice selection \\
\midrule
\multicolumn{4}{l}{\textbf{Generated-image comparison}}\\
Rapidata-4o~\citep{rapidata2026subjectivetruth} & RAP & Human judgments of generated images, including prompt alignment. & Pairwise image selection \\
GenAI-Bench~\citep{li2024genai} & GEN & Compositional text-to-image alignment and human judgments. & Pairwise image selection \\
\end{longtable}
\endgroup

\section{Question Category Annotation}
\label{supp:question_annotation}
Questions within the same benchmark can differ in the evidence needed
to answer them. Some can be resolved from text and prior knowledge,
whereas others require visual observations or the combination of
textual and visual evidence. To distinguish these requirements, we
assign each evaluated question to one of four categories:
text-sufficient (\textbf{T}), text-prior ($\mathbf{T^-}$),
visual-observation (\textbf{V}), or joint-vision and text (\textbf{J}).
These categories describe the evidence requirements inferred from
the question and its accompanying information. We first define
the taxonomy~\ref{appendix:evidence_taxonomy}, then describe its application~\ref{appendix:evidence_protocol} and validation~\ref{appendix:evidence_validation}, and
finally report the category distribution across benchmarks~\ref{appendix:evidence_composition}.

\subsection{Definition}
\label{appendix:evidence_taxonomy}

Each question receives exactly one category. The taxonomy distinguishes
whether the text determines the correct answer (\textbf{T}), merely suggests it ($\mathbf{T^-}$),
directs the reader to a visual observation (\textbf{V}), or supplies substantive
information that must be combined with the image (\textbf{J}).

\begin{description}
    \item[\textbf{T: Text sufficient.}]
    The question, answer options, and any supplied hint or context,
    together with general or domain knowledge, determine the correct
    option without the image. Examples include factual or definitional
    questions, option sets with only one logically possible answer,
    and hints that explicitly provide the answer. Operationally, a
    reader given the text alone would confidently select the gold
    option.

    \item[$\mathbf{T^-}$: Text prior.]
    The text makes the ground truth more plausible than the alternatives,
    but does not establish it with confidence: the particular image
    could change the answer. Examples include generic cause--effect
    questions about food webs and relations suggested by common sense
    but not guaranteed by the text. A reader given the text alone would
    lean toward the ground truth without being certain. This category
    is denoted A in the original annotation files; we use
    $\mathbf{T^-}$ throughout the paper.

    \item[\textbf{V: Visual observation.}]
    The answer depends on observing the image, while the text primarily
    specifies what to inspect. Examples include object identification,
    counting, localization, color recognition, reading a diagram label,
    and questions about whether an object is present. Such questions
    remain \textbf{V} even though their instructions are expressed
    in text.

    \item[\textbf{J: Joint evidence.}]
    The text supplies substantive content, such as a description,
    premise, or claim, that must be matched against or applied to
    the image. Neither modality alone determines the answer.
    Examples include selecting the image that best matches a
    description, checking a stated value or ranking against a
    chart, and interpreting text--image incongruity.
\end{description}

\paragraph{Benchmark-specific definition.}
The same definitions apply across benchmarks. The annotation
instructions additionally clarify boundaries for formats that can
otherwise be ambiguous:
\begin{itemize}
    \item \emph{AI2D~\citep{kembhavi2016diagram}}.
    Science facts independent of the particular diagram are
    \textbf{T}; labels, depicted stages, and connections that must
    be read from the diagram are \textbf{V}; generic cause-effect
    questions about a food web can be $\mathbf{T^-}$ when prior
    knowledge suggests but does not determine the answer.

    \item \emph{A-OKVQA~\citep{schwenk2022okvqa}}.
    A factual question about an object explicitly named in the text
    is \textbf{T} if the photograph is unnecessary.
    A common-sense answer that the photograph could overturn is
    $\mathbf{T^-}$, while an answer requiring observation of the
    photograph is \textbf{V}.

    \item \emph{Reefknot~\citep{zheng2025reefknot}}.
    Relations largely determined or suggested by the named entities
    and answer options can be \textbf{T} or $\mathbf{T^-}$.
    Spatial and action relations that depend on the depicted scene
    are \textbf{V}.

    \item \emph{NaturalBench~\citep{li2024naturalbench}.}
    Simple perception questions are \textbf{V}. Compositional
    descriptions that supply constraints to be checked against the
    image, including combinations of attributes, counts, relations,
    or negations, are \textbf{J}.

    \item \emph{HallusionBench~\citep{guan2024hallusionbench} and MME~\citep{fu2026mme}.}
    Questions about plain image content are \textbf{V}, whereas
    substantive claims involving a value, date, ranking, name,
    sentence, or translation to be verified against the image
    are \textbf{J}. A question is \textbf{T} only when the text
    alone establishes the gold answer. In particular, because
    HallusionBench can contain edited images that conflict with
    prior knowledge, a familiar answer is not sufficient for
    assignment to \textbf{T}.
\end{itemize}

\subsection{Annotation Protocol}
\label{appendix:evidence_protocol}

\paragraph{Annotated item pool.}
We assign evidence-requirement categories to all $11{,}372$
items in the intervention evaluation pool across $14$ benchmarks.
For each benchmark, we include up to the first $700$ valid
items. This is for limiting differences in benchmark contributions since HallusionBench~\citep{guan2024hallusionbench} and RealWorldQA~\citep{realworldqa2024} only contribute $447$ and $425$
valid items, respectively.
We include only items with at least one image, at least two answer options, and a ground-truth answer corresponding
to one of the provided options.

\paragraph{Information available for annotation.}
We use Claude Opus 5~\citep{anthropic2026claude} as an annotator. Annotations use the question, answer options, any supplied hint
or context passage, the ground truth, the benchmark's category
metadata, and the number of images. The annotator does not see
the images, model predictions, or any accuracy measurements. The ground truth is included to assess whether textual evidence
determines or merely favors the correct answer, particularly
at the \textbf{T}/$\mathbf{T^-}$ boundary. Each benchmark was processed in one annotation run supplied
with the category definitions, the relevant benchmark-specific
clarifications, and the full item list. The annotator was required
to return exactly one category for each item. All output files were checked for complete coverage, with one
label per item.

\subsection{Annotation Validation}
\label{appendix:evidence_validation}

We assess the annotations in two ways: agreement between two runs and correspondence with model behavior
when visual input is removed.

\paragraph{Judge agreement.}
For the $100$ AI2D~\citep{kembhavi2016diagram} data shared across models in the blank-image
evaluation, we compare the full annotation with a pilot annotation
produced before it. The two runs agree on $87\%$ of items, with Cohen's
$\kappa=0.80$~\citep{cohen1960coefficient}. Disagreements only occur at the textual-sufficiency
boundary (whether the question is \textbf{T} or $\mathbf{T^-}$).

\paragraph{Image removal validation.}
On the same $100$ AI2D~\citep{kembhavi2016diagram} data, models achieve
$89.1\%$ accuracy on \textbf{T} items with a blank image,
compared with $29.2\%$ on \textbf{V} items. Note that $29.2\%$ is close to the
four-option chance level of $25\%$.
This difference is consistent with the intended distinction
between textual sufficiency and visual dependence.

Additionally, we measure the accuracy drop by replacing actual image with blank image. As shown in Table~\ref{tab:e-removal-types}, image removal reduces accuracy by
$3.9$ percentage points on \textbf{T}, but $35.4$ on \textbf{V} and $28.9$ on
\textbf{J}.
The small reduction for \textbf{T} and larger reductions for
\textbf{V} and \textbf{J} support the intended distinctions.

\begin{table}[!htbp]
\centering
\small
\caption{
Accuracy reduction after replacing images with blank inputs,
aggregated over model-item pairs.
Larger reductions indicate greater sensitivity to visual
input removal.
}
\label{tab:e-removal-types}
\begin{tabular}{lrr}
\toprule
Category & Accuracy drop (pp) & Difference from \textbf{T} (pp) \\
\midrule
\textbf{T}       & 3.9  & 0.0  \\
$\mathbf{T^-}$   & 19.3 & 15.4 \\
\textbf{V}       & 35.4 & 31.5 \\
\textbf{J}       & 28.9 & 25.0 \\
\bottomrule
\end{tabular}
\end{table}

\subsection{Annotation Results}
\label{appendix:evidence_composition}

Table~\ref{tab:categories} reports the category counts for each
benchmark. The distribution varies substantially across benchmarks.
POPE~\citep{li2023evaluating} and CV-Bench~\citep{tong2024cambrian} are entirely \textbf{V}, while
GenAI-Bench~\citep{li2024genai} and Rapidata-4o~\citep{rapidata2026subjectivetruth} are entirely \textbf{J}.
In contrast, Reefknot~\citep{zheng2025reefknot} contains predominantly \textbf{T}
and $\mathbf{T^-}$, and AI2D~\citep{kembhavi2016diagram} and ScienceQA~\citep{lu2022learn} contain
substantial text-sufficient subsets.

A pooled comparison
between categories can also reflect differences between
benchmarks. We therefore use within-model-benchmark contrasts
for category comparisons intended to separate evidence
requirements from benchmark effects. Only benchmarks with
at least two represented categories can contribute to such
within-cell contrasts, and any additional group-size and
answer-balancing criteria are applied separately for the
corresponding information analysis. The counts below describe
the full annotation pool.

\begin{table}[!htbp]
\centering
\small
\setlength{\tabcolsep}{5pt}
\caption{
Evidence-requirement categories in the evaluated item pool.
}
\label{tab:categories}
\begin{tabular}{llrrrr}
\toprule
Benchmark  & Items
& \textbf{T} & $\mathbf{T^-}$ & \textbf{V} & \textbf{J}\\
\midrule
POPE~\citep{li2023evaluating}           &  $700$ & $0$ & $0$ & $700$ & $0$  \\
CV-Bench~\citep{tong2024cambrian}       &  $700$ & $0$ & $0$ & $700$ & $0$  \\
RealWorldQA~\citep{realworldqa2024}    &  $425$ & $0$ & $0$ & $425$ & $0$  \\
GenAI-Bench~\citep{li2024genai}    &  $700$ & $0$ & $0$ & $0$ & $700$  \\
Rapidata-4o~\citep{rapidata2026subjectivetruth}    &  $700$ & $0$ & $0$ & $0$ & $700$  \\
HallusionBench~\citep{guan2024hallusionbench} & $447$ & $123$ & $32$ & $199$ & $93$ \\
MME~\citep{fu2026mme}            & $700$ & $0$ & $0$ & $238$ & $462$  \\
NaturalBench~\citep{li2024naturalbench}   & $700$ & $0$ & $0$ & $480$ & $220$  \\
MMBench~\citep{liu2024mmbench}        & $700$ & $71$ & $24$ & $600$ & $5$ \\
SEED-Bench~\citep{li2023seed}     & $700$ & $7$ & $28$ & $645$ & $20$  \\
AI2D~\citep{kembhavi2016diagram}           & $700$ & $261$ & $124$ & $315$ & $0$ \\
A-OKVQA~\citep{schwenk2022okvqa} & $700$ & $69$ & $62$ & $569$ & $0$ \\
ScienceQA~\citep{lu2022learn}      & $700$ & $406$ & $21$ & $233$ & $40$  \\
Reefknot~\citep{zheng2025reefknot}       & $700$ & $481$ & $164$ & $55$ & $0$ \\
\midrule
Total & &  $1{,}676$ & $571$ & $6{,}744$ & $2{,}381$  \\
\bottomrule
\end{tabular}
\end{table}

\section{More Analysis}
\label{sec:detailed-method}
\subsection{Eq.~\ref{eq:modality_bases}, ~\ref{eq:unique_geometry}, ~\ref{eq:synergy_pca}: Determining \texorpdfstring{$r_b, r_u, r_s$}{r} in Batch-wise PCA}
\label{app:md-rank-r}
\begin{table}[t]
\centering
\small
\setlength{\tabcolsep}{5pt}
\caption{
Sensitivity of $U_v$ to the vision-unique rank cap $r_u$
across 6 models and 3 benchmarks.
Mean $U_v$ is the equally weighted average of the two
previously reported group means.
$\rho$ is the Spearman correlation of the model ordering
by $U_v$ with the default setting $r_u=32$.
The magnitude increases with the cap, while the model
ordering remains unchanged.
}
\label{tab:rank-diagnosis}
\begin{tabular}{rcc}
\toprule
$r_u$ & Mean $U_v$ & $\rho$ vs $r_u=32$ \\
\midrule
$16$  & $0.01455$ & $1.000$ \\
$32$  & $0.01890$ & $1.000$ \\
$64$  & $0.02495$ & $1.000$ \\
$128$ & $0.03115$ & $1.000$ \\
$256$ & $0.03965$ & $1.000$ \\
$512$ & $0.04785$ & $1.000$ \\
\bottomrule
\end{tabular}
\end{table}
\begin{table}[t]\centering\small\setlength{\tabcolsep}{5pt}
\caption{Accuracy gain by changing the rank of $\Guv$ during the intervention stage. Each entry is the mean accuracy gain (pp) over the six configurations the rank sweep and the main grid share ($\ell/L\in\{0.5,0.6,0.7\}\times\alpha\in\{2,4\}$), with the change from the $r_u{=}32$ default in parentheses.}
\label{tab:rank-intervention}
\begin{tabular}{lcccc}
\toprule
Model & $r_u=4$ & $r_u=8$ & $r_u=16$ & $r_u=32$ \\
\midrule
ViGoRL-7B~\citep{sarch2026grounded} & $+13.3$ ($-3.5$)  & $+21.1$ ($+4.2$)  & $+19.1$ ($+2.3$)  & $+16.9$  \\
ViGoRL-3B~\citep{sarch2026grounded} & $+2.6$ ($-0.9$)  & $+5.7$ ($+2.2$)  & $+3.9$ ($+0.4$)  & $+3.5$  \\
ViGoRL-MCTS-3B~\citep{sarch2026grounded} & $+2.4$ ($+0.0$)  & $+3.6$ ($+1.2$)  & $+3.4$ ($+1.0$)  & $+2.4$  \\
OpenVLThinker-7B~\citep{deng2025openvlthinker} & $-2.4$ ($+0.1$)  & $-2.9$ ($-0.4$)  & $-3.5$ ($-1.0$)  & $-2.5$  \\
Vision-R1-7B~\citep{huang2026vision} & $-3.2$ ($-0.2$)  & $-3.1$ ($-0.0$)  & $-2.7$ ($+0.3$)  & $-3.0$  \\

\bottomrule
\end{tabular}
\end{table}

The modality rank cap $r_b$ in Eq.~\ref{eq:modality_bases} controls the estimated subspace
removed from the opposite modality, while the unique rank cap
$r_u$ in Eq.~\ref{eq:unique_geometry} controls how many principal directions are retained
after this removal. Since modality representations contain both shared and unique
structure, we allocate a larger rank budget to the modality
bases ($r_b=64$) and a smaller budget to the unique directions
retained after removing the opposite modality's estimated
subspace ($r_u=32$). For the combined basis $B_{VT}$ in Eq.~\ref{eq:synergy_pca},
we allocate the sum of the two modality rank budgets,
$r_s=64+64=128$, to represent both modalities.

We evaluate rank sensitivity by varying $r_u$ from $16$
to $512$ while keeping $r_b$ and $r_s$ fixed
(Table~\ref{tab:rank-diagnosis}).
Although the magnitude of $U_v$ changes, the evaluated
models retain the same ordering (Spearman $\rho=1.000$
relative to $r_u=32$), supporting the stability of
cross-model comparisons over the tested range.
For intervention, rank $8$ improves gain over rank $32$
by a median $0.67$ percentage points across $40$ matched
model-benchmark pairs
(Table~\ref{tab:rank-intervention}).
This improvement at $r_u=8$ suggests separately calibrating the intervention rank may yield even further gains for the future work.

\begin{figure}[!ht]
\centering
\includegraphics[
    width=\linewidth,
]{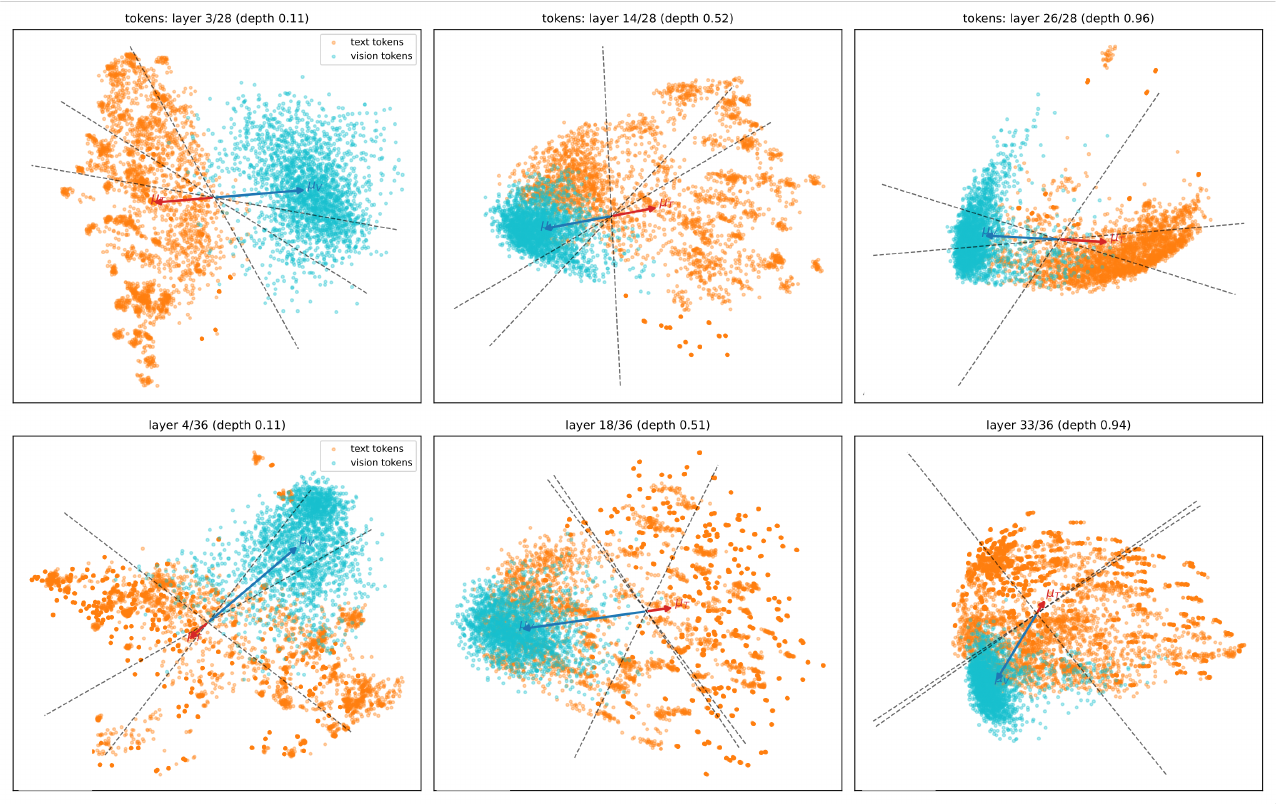}
\caption{
Illustrative two-dimensional projections of visual and
textual representations across decoder depth on MMBench~\citep{liu2024mmbench}.
Top: Qwen2.5-VL-7B~\citep{bai2025qwen25vltechnicalreport} at layers $3$, $14$, and $26$
under the displayed $28$-layer indexing.
Bottom: ViGoRL-3B~\citep{sarch2026grounded} at layers $4$, $18$, and $33$
under the displayed $36$-layer indexing.
\textcolor{cyan}{Cyan} and \textcolor{orange}{orange} points denote \textcolor{cyan}{visual} and \textcolor{orange}{textual}
representations, respectively. Arrows mark modality means.
Dashed lines indicate answer-readout axes.
}
\label{fig:e-token-geometry}
\end{figure}

\subsection{Details of Section~\ref{subsec:pid_diagnosis}}
\label{app:md-subspaces}

This subsection supplements the geometric construction in
Section~\ref{subsec:pid_diagnosis}.
We first visualize the modality representations used to
estimate the bases in Eq.~\ref{eq:modality_bases},
then establish the orthogonality of the unique directions
in Eq.~\ref{eq:unique_geometry}, and finally describe
how to implement the complementary projection defined
by Eq.~\ref{eq:synergy_geometry}.

\paragraph{Details of Eq.~\ref{eq:modality_bases}.}
Figure~\ref{fig:e-token-geometry} provides illustrative
two-dimensional projections of the vision and text
representations used to estimate the modality bases.
For both Qwen2.5-VL-7B~\citep{bai2025qwen25vltechnicalreport} and ViGoRL-3B~\citep{sarch2026grounded}, the relative
orientation and spread of the two representation
distributions vary across decoder depth.
In the late-layer, Qwen2.5-VL-7B~\citep{bai2025qwen25vltechnicalreport} retains
a broadly distributed visual representation cloud,
whereas ViGoRL-3B~\citep{sarch2026grounded} shows a more concentrated visual
distribution relative to its text representations.
These views illustrate differences in the underlying
representations.

\paragraph{Details of Eq.~\ref{eq:unique_geometry}.}
The unique bases are orthogonal to the opposite
modality's estimated subspace:
\begin{equation}
B_t^\top G_{U_v}=0,
\qquad
B_v^\top G_{U_t}=0.
\end{equation}
To see this, since $B_t$ has orthonormal columns,
\begin{equation}
B_t^\top(I-B_tB_t^\top)=0.
\end{equation}
Thus, the range of the residual covariance
$(I-B_tB_t^\top)C_v(I-B_tB_t^\top)$
is orthogonal to $B_t$.
Every eigenvector with a positive eigenvalue lies
in this range, so the retained directions forming
$G_{U_v}$ satisfy $B_t^\top G_{U_v}=0$.
The same argument applies to $G_{U_t}$ after
exchanging vision and text.

\paragraph{Details of Eq.~\ref{eq:synergy_geometry}.}
The complementary projection can be computed without
explicitly constructing
$G_S\in\mathbb{R}^{d\times(d-r_{VT})}$,
where $r_{VT}$ is the number of orthonormal columns
in $B_{VT}$.
Let $H$ contain the decision representations as rows.
Using
$G_SG_S^\top=I-B_{VT}B_{VT}^\top$,
we compute the complementary representations directly:
\begin{equation}
H(I-B_{VT}B_{VT}^\top)
=
H-(HB_{VT})B_{VT}^\top.
\end{equation}
This representation preserves the Gram matrix obtained
from explicit complementary coordinates:
\begin{align}
&\bigl[H(I-B_{VT}B_{VT}^\top)\bigr]
\bigl[H(I-B_{VT}B_{VT}^\top)\bigr]^\top
\nonumber\\
&\qquad=
H(I-B_{VT}B_{VT}^\top)H^\top
=
(HG_S)(HG_S)^\top.
\end{align}
Thus, both implementations yield identical pairwise
inner products and norms, without requiring storage
of an explicit complementary basis.

\subsection{Implementation of Information Measurement (Eq.~\ref{eq:renyi_mi})}
\label{app:md-info-proofs}

For the cosine kernel $K_Z$ with unit diagonal and
the label kernel $\mathbf{1}[y_i=y_j]$, we compute
the matrix-based R\'enyi-2 mutual information~\citep{giraldo2014measures,yu2019multivariate,yu2018multivariate} directly
from the kernel entries:
\begin{equation}
I_2(Z;Y)
=
\log
\frac{
N_{\mathcal B}^{\,2}
\displaystyle\sum_{\substack{i,j\in\mathcal B\\y_i=y_j}}
K_Z[i,j]^2
}{
\left(
\displaystyle\sum_{i,j\in\mathcal B}K_Z[i,j]^2
\right)
\left(\displaystyle\sum_c n_c^2\right)
},
\end{equation}
where $n_c$ is the number of examples with label $c$
in the batch.
In implementation, we square the entries of $K_Z$
and sum them over all sample pairs and over pairs
with matching labels.
Together with the label counts, these sums give
$I_2(Z;Y)$ without an eigendecomposition.
This expression is exactly equivalent to the
trace-normalized R\'enyi-2 formulation, since the
trace of a symmetric matrix squared equals the
sum of its squared entries.

\subsection{Validating the Correspondence Dependence (Eq.~\ref{eq:synergy}) of \texorpdfstring{$S$}{S}} \label{app:e-correspondence} 
Section~\ref{subsec:pid_diagnosis} defines $S$ as the difference in answer information measured within $G_S$ between matched and mismatched image--text inputs. To examine whether this difference depends on replacing the image with one from another category, we compare matched images, same-category substitutes, different-category substitutes, and blank inputs. Table~\ref{tab:e-shuffle-cells} reports three examples. For InternVL3-8B~\citep{zhu2025internvl3} on MMBench~\citep{liu2024mmbench}, answer information within $G_S$ falls from $0.4150$ with the matched image to $0.0323$ with a same-category substitute, close to $0.0274$ with a different-category substitute. The other two examples show the same pattern. Thus, the reduction occurs even when the substitute preserves the image category. This supports interpreting $S$ as sensitive to the specific image-text correspondence, rather than attributing the reduction solely to a change in image category or the removal of visual input. 
\begin{table}[!ht] 
\centering 
\small 
\caption{Answer information measured within $G_S$ under various image substitutions. Three representative examples are shown. `Same category' and `Different category' indicate whether the original image was replaced with another image from the same question category or from a completely different category.} 
\label{tab:e-shuffle-cells} 
\resizebox{\textwidth}{!}{%
\begin{tabular}{lrrrr} 
\toprule 
Model / Benchmark & Matched & Same category & Different category & Blank \\ 
\midrule 
InternVL3-8B~\citep{zhu2025internvl3} / MMBench~\citep{liu2024mmbench} & 0.4150 & 0.0323 & 0.0274 & 0.0149 \\ 
Qwen2.5-VL-7B~\citep{bai2025qwen25vltechnicalreport} / POPE~\citep{li2023evaluating} & 0.1882 & 0.0106 & 0.0055 & 0.0010 \\ 
ViGoRL-3B~\citep{sarch2026grounded} / MMBench~\citep{liu2024mmbench} & 0.0098 & 0.0033 & 0.0031 & 0.0019 \\ 
\bottomrule 
\end{tabular}%
}
\end{table}

\begin{table}[!ht]\centering\small
\caption{Information measurements across benchmarks.}
\label{tab:e-datasets}
\begin{tabular}{lrrrrr}
\toprule
Benchmark & $I_2$ & $R$ & $U_v$ & $U_t$ & $S$ \\
\midrule
MMBench~\citep{liu2024mmbench} & 0.0483 & 0.0063 & 0.0595 & 0.0566 & 0.3001 \\
A-OKVQA~\citep{schwenk2022okvqa} & 0.0501 & 0.0067 & 0.0435 & 0.0408 & 0.2973 \\
SEED-Bench~\citep{li2023seed} & 0.0340 & 0.0052 & 0.0367 & 0.0308 & 0.1940 \\
POPE~\citep{li2023evaluating} & 0.0459 & 0.0153 & 0.0386 & 0.0385 & 0.1860 \\
Reefknot~\citep{zheng2025reefknot} & 0.0341 & 0.0034 & 0.0275 & 0.0192 & 0.1844 \\
MME~\citep{fu2026mme} & 0.0349 & 0.0111 & 0.0358 & 0.0355 & 0.1738 \\
ScienceQA~\citep{lu2022learn} & 0.0516 & 0.0079 & 0.0651 & 0.0382 & 0.1641 \\
AI2D~\citep{kembhavi2016diagram} & 0.0368 & 0.0047 & 0.0446 & 0.0663 & 0.1634 \\
NaturalBench~\citep{li2024naturalbench} & 0.0145 & 0.0036 & 0.0159 & 0.0113 & 0.0796 \\
HallusionBench~\citep{guan2024hallusionbench} & 0.0122 & 0.0019 & 0.0136 & 0.0160 & 0.0787 \\
CV-Bench~\citep{tong2024cambrian} & 0.0227 & 0.0047 & 0.0188 & 0.0099 & 0.0652 \\
RealWorldQA~\citep{realworldqa2024} & 0.0131 & 0.0026 & 0.0115 & 0.0136 & 0.0543 \\
Rapidata-4o~\citep{rapidata2026subjectivetruth} & 0.0021 & 0.0008 & 0.0020 & 0.0009 & 0.0160 \\
GenAI-Bench~\citep{li2024genai} & 0.0019 & 0.0009 & 0.0020 & 0.0007 & 0.0088 \\
\bottomrule
\end{tabular}
\end{table}

\subsection{Figure~\ref{fig:layerwise}(b) Across All Benchmarks}
\label{app:e-benchmarks}

\begin{figure}[!htbp]\centering
\includegraphics[width=\linewidth]{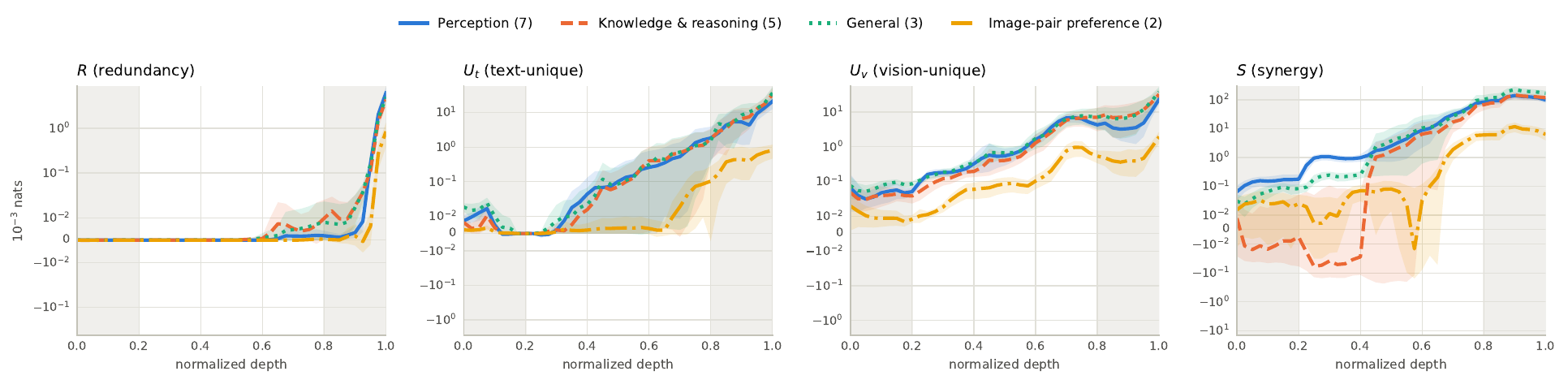}
\caption{Layer-wise information profiles by the tasks: perception ($7$ benchmarks), knowledge and reasoning ($5$), general ($3$), and image-pair preference ($2$). Image-pair preference tasks show a later rise in $U_t$ than others.}
\label{fig:e-task-profiles}
\end{figure}

\paragraph{Different tasks expose different profiles.}
Table~\ref{tab:e-datasets} reports the PID measurements across benchmarks. MMBench~\citep{liu2024mmbench} and A-OKVQA~\citep{schwenk2022okvqa} have the largest reported $S$, followed by SEED-Bench~\citep{li2023seed} and POPE~\citep{li2023evaluating}. Rapidata-4o~\citep{rapidata2026subjectivetruth} and GenAI-Bench~\citep{li2024genai} have much smaller absolute values, $0.0160$ and $0.0088$. Figure~\ref{fig:e-task-profiles} groups the benchmarks into perception, knowledge and reasoning, general, and image-pair preference tasks. Image-pair preference tasks show a later rise in $U_t$: it remains near zero through much of the middle layers, while the other task families show an earlier increase. Thus, task families differ not only in the amount of measured answer information but also in the depth at which it becomes apparent.

\paragraph{Overall information($I_2$) is different from image-text correspondence information ($S$).}
Table~\ref{tab:e-paired} compares Qwen2.5-VL-7B~\citep{bai2025qwen25vltechnicalreport} and Vision-R1-7B~\citep{huang2026vision}. Qwen2.5-VL-7B~\citep{bai2025qwen25vltechnicalreport} has higher $S$ on every benchmark, with gaps ranging from $0.011$ on GenAI-Bench~\citep{li2024genai} to $0.265$ on MMBench~\citep{liu2024mmbench}. Notably, while Qwen2.5-VL-7B extracts nearly identical overall answer information ($I_2 \approx 0.049 - 0.050$) across ScienceQA~\citep{lu2022learn}, MME~\citep{fu2026mme}, and A-OKVQA~\citep{schwenk2022okvqa}, its synergy ($S$) varies significantly across these tasks (0.163, 0.201, and 0.289, respectively). This demonstrates that similar overall answer information can coexist with vastly different amounts of correspondence-dependent information, depending on the task's specific visual requirements.

\begin{table}[!htbp]\centering\small
\caption{Per-benchmark information comparison for Qwen2.5-VL-7B~\citep{bai2025qwen25vltechnicalreport} (Q) and Vision-R1-7B~\citep{huang2026vision} (V1). $\Delta S=S_{\mathrm Q}-S_{\mathrm{V1}}$ is calculated from the rounded entries.}
\label{tab:e-paired}
\begin{tabular}{lrrrrr}
\toprule
Benchmark & $S_{\mathrm Q}$ & $S_{\mathrm{V1}}$ & $\Delta S$ & $I_{2,\mathrm Q}$ & $I_{2,\mathrm{V1}}$ \\
\midrule
POPE~\cite{li2023evaluating} & 0.183 & 0.036 & 0.147 & 0.064 & 0.005 \\
MME~\citep{fu2026mme} & 0.201 & 0.061 & 0.140 & 0.050 & 0.009 \\
MMBench~\citep{liu2024mmbench} & 0.309 & 0.044 & 0.265 & 0.047 & 0.006 \\
HallusionBench~\cite{guan2024hallusionbench} & 0.109 & 0.016 & 0.093 & 0.015 & 0.003 \\
SEED-Bench~\citep{li2023seed} & 0.166 & 0.045 & 0.121 & 0.029 & 0.006 \\
AI2D~\citep{kembhavi2016diagram} & 0.200 & 0.051 & 0.149 & 0.043 & 0.010 \\
CV-Bench~\citep{tong2024cambrian} & 0.070 & 0.017 & 0.053 & 0.020 & 0.004 \\
Rapidata-4o~\citep{rapidata2026subjectivetruth} & 0.028 & 0.006 & 0.022 & 0.004 & 0.000 \\
GenAI-Bench~\citep{li2024genai} & 0.014 & 0.003 & 0.011 & 0.003 & 0.001 \\
ScienceQA~\citep{lu2022learn} & 0.163 & -0.003 & 0.166 & 0.050 & 0.004 \\
A-OKVQA~\citep{schwenk2022okvqa} & 0.289 & 0.037 & 0.252 & 0.049 & 0.005 \\
NaturalBench~\citep{li2024naturalbench} & 0.094 & 0.020 & 0.074 & 0.019 & 0.003 \\
RealWorldQA~\citep{realworldqa2024} & 0.062 & 0.007 & 0.055 & 0.013 & 0.002 \\
Reefknot~\citep{zheng2025reefknot} & 0.170 & 0.009 & 0.161 & 0.025 & 0.002 \\
\bottomrule
\end{tabular}
\end{table}

\providecommand{\EAnalysisFigsPath}{figure}
\providecommand{\EAnalysisFigurePath}{figure}

\subsection{Details of Table~\ref{tab:exp-evidence}:
How \texorpdfstring{$S$}{S} Relates to Visual Information Use}
\label{app:e-correlations}

\paragraph{Information measurements correlate with accuracy
and sensitivity to image removal.}
Table~\ref{tab:e-correlations} reports Spearman correlations
between the information measurements and model behavior. Across pooled model--benchmark pairs, $S$ correlates
with image-removal accuracy drop at $0.73$ and with
base accuracy at $0.83$, compared with $0.70$ and $0.82$
for $I_2(H;Y)$.

\begin{table}[!ht]\centering\small
\caption{Spearman correlations between information measurements and image-removal accuracy drop, with pooled rows for removal drop and base accuracy. The pooled results are reported for $308$ cells.}
\label{tab:e-correlations}
\begin{tabular}{lrrrrrr}
\toprule
Benchmark / outcome & $R$ & $U_v$ & $U_t$ & $S$ & $I_2$ & $R-S$ \\
\midrule
POPE~\citep{li2023evaluating} & +0.14 & +0.59 & +0.40 & +0.46 & +0.50 & -0.58 \\
MME~\citep{fu2026mme} & +0.77 & +0.31 & +0.84 & +0.57 & +0.75 & -0.56 \\
MMBench~\citep{liu2024mmbench} & +0.53 & +0.25 & +0.57 & +0.65 & +0.63 & -0.64 \\
HallusionBench~\citep{guan2024hallusionbench} & +0.50 & +0.37 & +0.83 & +0.80 & +0.63 & -0.80 \\
Rapidata-4o~\citep{rapidata2026subjectivetruth} & +0.07 & +0.26 & +0.25 & +0.33 & +0.30 & -0.36 \\
GenAI-Bench~\citep{li2024genai} & +0.81 & +0.57 & +0.90 & +0.83 & +0.89 & -0.77 \\
SEED-Bench~\citep{li2023seed} & +0.53 & +0.68 & +0.44 & +0.83 & +0.79 & -0.84 \\
AI2D~\citep{kembhavi2016diagram} & +0.69 & +0.47 & +0.58 & +0.91 & +0.67 & -0.91 \\
ScienceQA~\citep{lu2022learn} & +0.43 & +0.60 & +0.48 & +0.65 & +0.65 & -0.64 \\
A-OKVQA~\citep{schwenk2022okvqa} & +0.53 & +0.22 & +0.63 & +0.45 & +0.56 & -0.44 \\
CV-Bench~\citep{tong2024cambrian} & +0.67 & +0.33 & +0.80 & +0.83 & +0.80 & -0.82 \\
NaturalBench~\citep{li2024naturalbench} & +0.78 & +0.52 & +0.90 & +0.66 & +0.84 & -0.65 \\
RealWorldQA~\citep{realworldqa2024} & +0.67 & +0.43 & +0.80 & +0.78 & +0.76 & -0.77 \\
Reefknot~\citep{zheng2025reefknot} & +0.32 & +0.13 & -0.09 & +0.13 & +0.06 & -0.12 \\
Pooled: removal drop & +0.62 & +0.63 & +0.66 & +0.73 & +0.70 & -0.73 \\
Pooled: base accuracy & +0.67 & +0.70 & +0.75 & +0.83 & +0.82 & -0.83 \\
\bottomrule
\end{tabular}
\end{table}

\paragraph{Sensitivity to excluding one model family.}
To test whether the association between $S$ and
image-removal accuracy drop is driven by a single
model family, we repeat the analysis after excluding
its three models (Table~\ref{tab:e-family-sensitivity}).
The correlation remains positive at $0.68$ across
models and $0.67$ across model--benchmark pairs,
compared with $0.79$ and $0.73$ in the full analysis.

\begin{table}[!htbp]\centering\small
\caption{Sensitivity of the $S$: image-removal association to excluding the ViGoRL~\citep{sarch2026grounded} family.}
\label{tab:e-family-sensitivity}
\begin{tabular}{llrr}\toprule
Aggregation & Included models & Count & Spearman $\rho$\\\midrule
Model-level & All & 22 & 0.79\\
Model-level & Excluding ViGoRL~\citep{sarch2026grounded} & 20 & 0.68\\
Cell-level & All & 308 & 0.73\\
Cell-level & Excluding ViGoRL~\citep{sarch2026grounded} & 340 & 0.67\\\bottomrule
\end{tabular}\end{table}

\subsection{Details of Figure~\ref{fig:item-guv}}
\label{app:e-correctness}

We test whether the higher vision-unique
geometry ($G_{U_v}$) for correct \textbf{V}/\textbf{J}
 than for correct \textbf{T}  persists under
different benchmark inclusion criteria.
Across the eight benchmarks with at least ten correct
\textbf{T} , the mean standardized
$G_{U_v}(\text{correct V/J})-G_{U_v}(\text{correct T})$
is $+0.71$.
Correct \textbf{V}/\textbf{J} have higher
$G_{U_v}$ than correct \textbf{T} in all eight
benchmarks. Including two additional benchmarks with only two correct
\textbf{T}  each reduces the mean standardized
$G_{U_v}(\text{correct V/J})-G_{U_v}(\text{correct T})$
to $+0.45$, still showing higher
$G_{U_v}$ than correct \textbf{T}  in nine of ten
benchmarks (Table~\ref{tab:e-pilot-inference}).
Tests comparing $G_{U_v}$ between correct
\textbf{V}/\textbf{J} and correct \textbf{T} 
yield $p=0.0002$ under both inclusion criteria,
using $4{,}000$ within-benchmark type-label permutations.
Thus, the higher $G_{U_v}$ for correct
\textbf{V}/\textbf{J}  remains evident even when
benchmarks with sparse correct-\textbf{T} samples
are included.

\begin{table}[!htbp]
\centering
\small
\setlength{\tabcolsep}{4pt}
\caption{
Robustness of higher $G_{U_v}$ for correct
\textbf{V}/\textbf{J}  than for correct \textbf{T}
 to benchmark inclusion criteria.
The second column reports the mean standardized
$G_{U_v}(\text{correct V/J})-G_{U_v}(\text{correct T})$;
the third gives its benchmark-bootstrap $95\%$
confidence interval.
Permutation tests compare these two sets of correct
items by shuffling type labels within benchmarks.
The final column counts benchmarks where correct
\textbf{V}/\textbf{J}  have higher $G_{U_v}$
than correct \textbf{T} .
}
\label{tab:e-pilot-inference}
\begin{tabular}{lcccc}
\toprule
Benchmark inclusion
& \shortstack{Mean standardized $G_{U_v}$\\
(correct V/J $-$ correct T)}
& $95\%$ CI
& $p$
& \shortstack{Benchmarks with higher $G_{U_v}$\\
for correct V/J than correct T} \\
\midrule
$\ge10$ correct \textbf{T}
& $+0.71$ & $[+0.46,+0.99]$ & $0.0002$ & $8/8$ \\
All with both types
& $+0.45$ & $[+0.005,+0.82]$ & $0.0002$ & $9/10$ \\
\bottomrule
\end{tabular}
\end{table}

\subsection{Details of Eq.~\ref{eq:vision_intervention}}
\label{app:md-intervention}

We describe the properties and implementation of
Eq.~\ref{eq:vision_intervention}.
All identities below apply at the intervention layer
$\ell^\star$, where $G_{U_v}^{\ell^\star}$ has
orthonormal columns.

\paragraph{Selective amplification.}
The intervention multiplies the components along
$G_{U_v}^{\ell^\star}$ by $1+\alpha$ and leaves
the orthogonal components unchanged:
\begin{align}
V_i^{\prime\,\ell^\star}G_{U_v}^{\ell^\star}
&=
(1+\alpha)V_i^{\ell^\star}G_{U_v}^{\ell^\star},
\\
V_i^{\prime\,\ell^\star}
\left(I-G_{U_v}^{\ell^\star}
(G_{U_v}^{\ell^\star})^\top\right)
&=
V_i^{\ell^\star}
\left(I-G_{U_v}^{\ell^\star}
(G_{U_v}^{\ell^\star})^\top\right).
\end{align}
The orthogonality established in
Eq.~\ref{eq:unique_geometry} also gives
\begin{equation}
V_i^{\prime\,\ell^\star}B_t^{\ell^\star}
=
V_i^{\ell^\star}B_t^{\ell^\star}.
\end{equation}
Thus, the intervention preserves the coordinates along the
estimated text basis at the intervention layer.

\paragraph{Invertibility.}
For $\alpha\geq0$, the intervention operator has
eigenvalue $1+\alpha$ on the vision-unique subspace
and $1$ on its orthogonal complement.
Its inverse is
\begin{equation}
\left(
I+\alpha G_{U_v}^{\ell^\star}
(G_{U_v}^{\ell^\star})^\top
\right)^{-1}
=
I-\frac{\alpha}{1+\alpha}
G_{U_v}^{\ell^\star}
(G_{U_v}^{\ell^\star})^\top.
\end{equation}
The intervention therefore preserves the matrix rank of
$V_i^{\ell^\star}$ and removes no representation
directions.

\paragraph{Efficient implementation.}
Instead of constructing a $d\times d$ projection
matrix, we apply the intervention as
\begin{equation}
V_i^{\prime\,\ell^\star}
=
V_i^{\ell^\star}
+
\alpha
\left(V_i^{\ell^\star}G_{U_v}^{\ell^\star}\right)
(G_{U_v}^{\ell^\star})^\top.
\end{equation}
For $n_i^v$ vision representations and rank cap $r_u$,
the matrix multiplications cost $O(n_i^v d r_u)$.

\subsection{Details of Section~\ref{sec:exp-rule}}
\label{supp:exp-rule-details}

To identify patterns that guide our intervention settings, we analyze the relationship between the base models' geometric profiles and the resulting accuracy gains. Table~\ref{tab:gain-correlation} presents the correlations across all 308 model-benchmark combinations. We observe that individual geometric shifts, such as a lower baseline synergy $S$, a larger late-layer decline in shared geometry ($\Delta G_R$), and a late-layer rise in text-unique geometry, are moderately predictive of intervention benefits. To assess the combined predictive power of these profiles, we fit gradient-boosted tree regressors~\citep{friedman2001greedy} using 10-fold cross-validation. While combining static layer differences yields a Pearson correlation of $0.66$, incorporating the full layer-wise trajectories of $G_R$, $G_{U_t}$, $G_{U_v}$, and $S$ across all layers increases the correlation to $r=0.80$. This strong predictive relationship demonstrates that the layer-wise evolution of representation geometry reliably indicates the potential benefit of vision-unique amplification, directly motivating the geometry-based intervention rules applied in Section~\ref{sec:exp-rule}.

\begin{table}[h]
\centering
\caption{Correlation between intervention gain of Table~\ref{tab:percell-qv-filtered} and the base model's geometry across 308 model $\times$ benchmark combinations. $\Delta G_R$ measures the drop in shared geometry (the peak value at depths $\geq0.5$ minus the output value). $\Delta (G_R$, $G_{U_t}$, $G_{U_v})$ indicates the late-layer rise in text-unique geometry minus the drops in both vision-unique and shared geometries. Boosted refers to out-of-fold predictions from gradient-boosted trees~\citep{friedman2001greedy} (10-fold cross-validation), from the three features $G_R$, $G_{U_t}$, $G_{U_v}$ or from the full layer-wise geometric trajectories.}
\label{tab:gain-correlation}
\resizebox{0.95\linewidth}{!}{%
\setlength{\tabcolsep}{4pt}%
\begin{tabular}{cccccc}
\toprule
$S$ &
\begin{tabular}[c]{@{}c@{}}$\Delta G_R$\end{tabular} &
$\Delta (G_R$, $G_{U_t}$, $G_{U_v})$ &
\begin{tabular}[c]{@{}c@{}}$\Delta (G_R$, $G_{U_t}$, $G_{U_v}$, $S)$\end{tabular} &
\begin{tabular}[c]{@{}c@{}}$\Delta (G_R$, $G_{U_t}$, $G_{U_v})$\\Boosted\end{tabular} &
\begin{tabular}[c]{@{}c@{}}full geometry,\\Boosted\end{tabular} \\
\midrule
\begin{tabular}[c]{@{}c@{}}$-0.54$\\(Spearman)\end{tabular} &
\begin{tabular}[c]{@{}c@{}}$0.54$\\(Pearson)\end{tabular} &
\begin{tabular}[c]{@{}c@{}}$0.56$\\(Pearson)\end{tabular} &
\begin{tabular}[c]{@{}c@{}}$0.61$\\(Pearson)\end{tabular} &
\begin{tabular}[c]{@{}c@{}}$0.66$\\(Pearson)\end{tabular} &
\begin{tabular}[c]{@{}c@{}}$0.80$\\(Pearson)\end{tabular} \\
\bottomrule
\end{tabular}}
\end{table}


\section{More Results}
\label{supp:results}
\providecommand{\FResultsFigsPath}{figure}
\providecommand{\FResultsFigurePath}{figure}



\begin{figure}[!ht]\centering
\includegraphics[width=\linewidth]{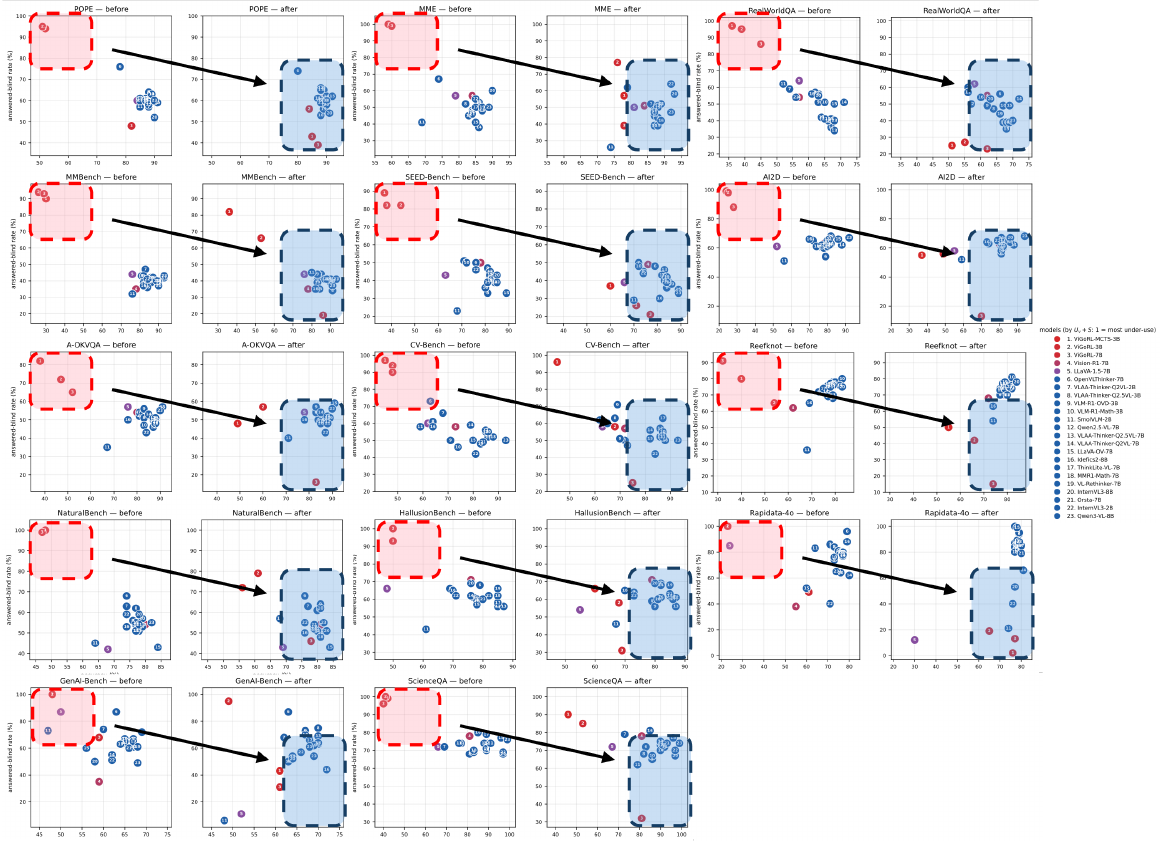}
\caption{Qualitative impact of the vision-unique ($G_{U_v}$) intervention across 14 benchmarks. The scatter plots illustrate the relationship between overall accuracy (x-axis) and the answered-blind rate (y-axis) before and after the intervention. Models identified as having strong text dominance, initially clustered in the \textcolor{red}{red} dashed boxes with high answered-blind rates, demonstrate a distinct shift towards the \textcolor{blue}{blue} dashed boxes following the intervention. This collective shift indicates a significant reduction in blind text reliance and a concurrent improvement in visual grounding and overall performance.}
\label{fig:qualitative}
\end{figure}
\subsection{Qualitative Results}
\label{app:qualitative}
\begin{figure}[!ht]\centering
\includegraphics[width=0.74\linewidth]{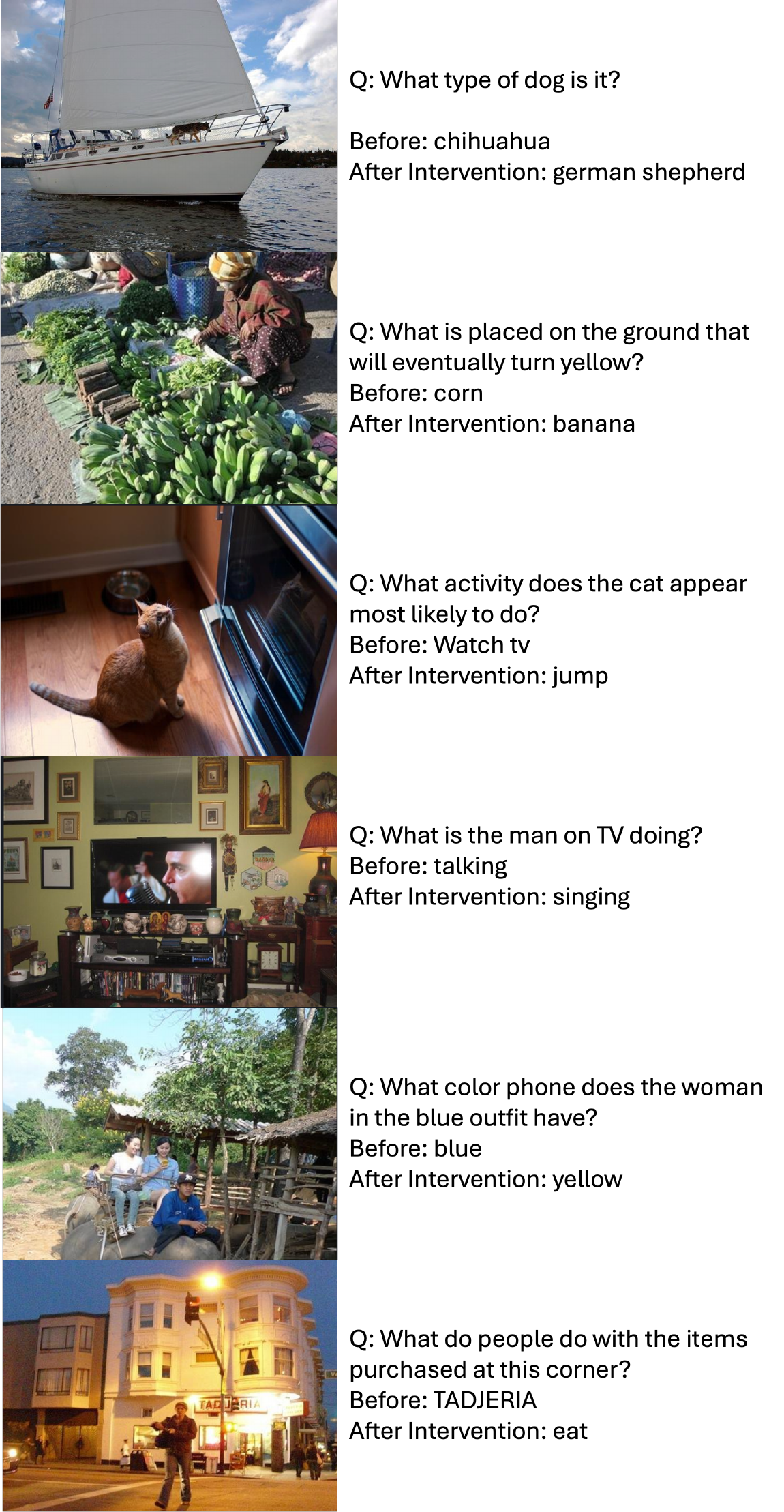}
\caption{Qualitative examples of open-ended generation before and after the $G_{U_v}$ intervention. Baseline models often fall back on language priors (e.g., blindly guessing ``corn" or ``chihuahua"), whereas the intervened models accurately ground their generated text in the specific visual evidence (e.g., ``banana", ``german shepherd").}
\label{fig:open-ended}
\end{figure}

Figure~\ref{fig:qualitative} demonstrates the qualitative effects of the selective amplification of the vision-unique subspace ($G_{U_v}$) on the predictive behavior of models across multiple benchmarks. Models as having weak vision-unique components and dominant text-unique components are located within the \textcolor{red}{red} dashed boxes in the `before' graphs, exhibiting a very high `answered-blind rate'. However, as a result of amplifying the representation components corresponding to the vision-unique geometric direction ($G_{U_v}$) through \textsc{GeoPID} (`after'), the distribution of these models shows a shift along the arrows into the \textcolor{blue}{blue} dashed box regions. This shift indicates that the rate at which models blindly rely on text to derive the answer has sharply decreased, while the overall performance metric (x-axis) has simultaneously improved. Consequently, this visualization strongly supports that the targeted intervention based on geometric structure activates synergy ($G_S$) in questions requiring both visual information and textual context, restoring the model's ability to utilize visual grounding.

\paragraph{Improvements in Open-Ended Generation.} The effectiveness of \textsc{GeoPID} extends beyond multiple-choice evaluation to open-ended visual question answering. As illustrated in Figure~\ref{fig:open-ended}, baseline VLMs frequently rely on language priors or exhibit text-biased hallucinations when generating free-form text. For instance, when asked to identify the green items placed on the ground that will eventually turn yellow, the baseline model incorrectly guesses ``corn" rather than grounding its answer in the visual evidence of unripe bananas. Similarly, it defaults to generic or stereotypical predictions like ``talking" or ``chihuahua" instead of accurately identifying the man singing on the TV or the German Shepherd on the boat. By selectively amplifying the vision-unique subspace ($G_{U_v}$), \textsc{GeoPID} effectively overrides these misleading text priors, enabling the model to generate accurate, visually grounded responses in open-ended settings.

\subsection{Accuracy Gain by Question Types}
\label{app:f-question-types}

\paragraph{Largest gains on joint-evidence questions.}
In the main paper, Table~\ref{tab:census-types-filtered} shows accuracy gain after the $G_{U_V}$ intervention for Table~\ref{tab:percell-qv-filtered}. Table~\ref{tab:f-category-transfer} further reports accuracy gains after the $G_{U_V}$ intervention for Table~\ref{tab:percell-geotree_d4-filtered}, for the four question types defined in
Appendix~\ref{supp:question_annotation}.
Joint-evidence questions (\textbf{J}), which require
combining textual and visual evidence, show the largest
gain.
This pattern is consistent with the intended role of
\textsc{GeoPID} in improving predictions that require
visual evidence alongside text.


\begin{table}[!htbp]
\centering
\small
\caption{
Accuracy gain by question category ($\%$).
}
\label{tab:f-category-transfer}
\begin{tabular}{rrrrr}
\toprule
$\mathbf{T}$ & $\mathbf{T^-}$ & $\mathbf{V}$ & $\mathbf{J}$ & All \\
\midrule
$+3.20$ & $+2.90$ & $+3.15$ & $+4.40$ & $+3.40$ \\
\bottomrule
\end{tabular}
\end{table}


\paragraph{Gains on visually dependent questions require the matching image.}
Table~\ref{tab:f-image-substitution} compares intervention
gains with the original image, an image reassigned from
another question, and a blank image.
For \textbf{V} and \textbf{J} questions, gains with the
original image are $13.1$ and $13.9$ percentage points,
respectively.
With mismatched images, these gains fall to $1.0$ and
$-3.3$ points, while blank images yield little change.
Thus, the gains on visually dependent questions rely
on the image matching the question, not because of merely the presence of an image.
For \textbf{T} questions, gains remain substantial with
mismatched images, indicating that some improvements
can also occur without the original visual evidence.

\begin{table}[!htbp]\centering\small
\caption{Image-condition control on the $100$ data for the models exhibiting strong
text dominance in Figure~\ref{fig:rule}. We report the accuracy gains in $\%$.}
\label{tab:f-image-substitution}
\begin{tabular}{lrrr}\toprule
Question category & True image & Deranged image & Blank image\\\midrule
$T$: text decides & $+22.6$ & $+17.7$ & $+1.3$\\
$A$: text prior & $+15.3$ & $+7.4$ & $0.0$\\
$V$: image reading & $+13.1$ & $+1.0$ & $-0.2$\\
$J$: joint evidence & $+13.9$ & $-3.3$ & $+0.3$\\\bottomrule
\end{tabular}
\end{table}

\subsection{Understanding the Accuracy Gains}
\label{app:f-collapse}

We further examine the prediction changes underlying
the accuracy gains in
Tables~\ref{tab:percell-qv-filtered}
and~\ref{tab:percell-geotree_d4-filtered}.

\paragraph{Reduced concentration on a single answer option.}
We examine how often the models exhibiting strong
text dominance in Figure~\ref{fig:rule} select the same answer option across questions.
As shown in Table~\ref{tab:f-collapse}, the most frequently
selected option accounts for $72\%$ of predictions before
intervention and $54\%$ afterward.
For comparison, the most frequent ground-truth option
accounts for $39\%$ of answers.
The intervention therefore reduces the tendency to
repeatedly select a single option.

\begin{table}[!htbp]
\centering
\small
\caption{
Concentration of predictions on the most frequently
selected answer option before and after intervention.
The ground-truth majority share is provided for comparison.
}
\label{tab:f-collapse}
\begin{tabular}{lr}
\toprule
Statistic & Share \\
\midrule
Most frequent predicted option before intervention & $72\%$ \\
Most frequent predicted option after intervention & $54\%$ \\
Most frequent ground-truth option & $39\%$ \\
\bottomrule
\end{tabular}
\end{table}

\paragraph{Changed answers are more often correct than random switches.}
We next examine whether initially incorrect predictions
become correct when the intervention changes the answer.
We consider the $678$ model-item pairs with at least
three answer options, an initially incorrect prediction,
and a different prediction after intervention.
Among these pairs, $74.0\%$ of the new predictions are
correct (Table~\ref{tab:f-conditional-correction}).
Uniformly selecting an option other than the initial
incorrect answer would yield an expected accuracy of
$34.1\%$.
Thus, the changed predictions favor the correct answer
more often than random switching would.

\begin{table}[!htbp]
\centering
\small
\caption{
Conditional correction among initially incorrect
predictions that change after intervention.
Only model-item pairs with at least three answer
options are included.
}
\label{tab:f-conditional-correction}
\begin{tabular}{lr}
\toprule
Statistic & Value \\
\midrule
Eligible model--item pairs & $678$ \\
Correct after intervention & $74.0\%$ \\
Expected correctness under random switching & $34.1\%$ \\
\bottomrule
\end{tabular}
\end{table}

\begin{figure}[!ht]\centering
\includegraphics[width=\linewidth]{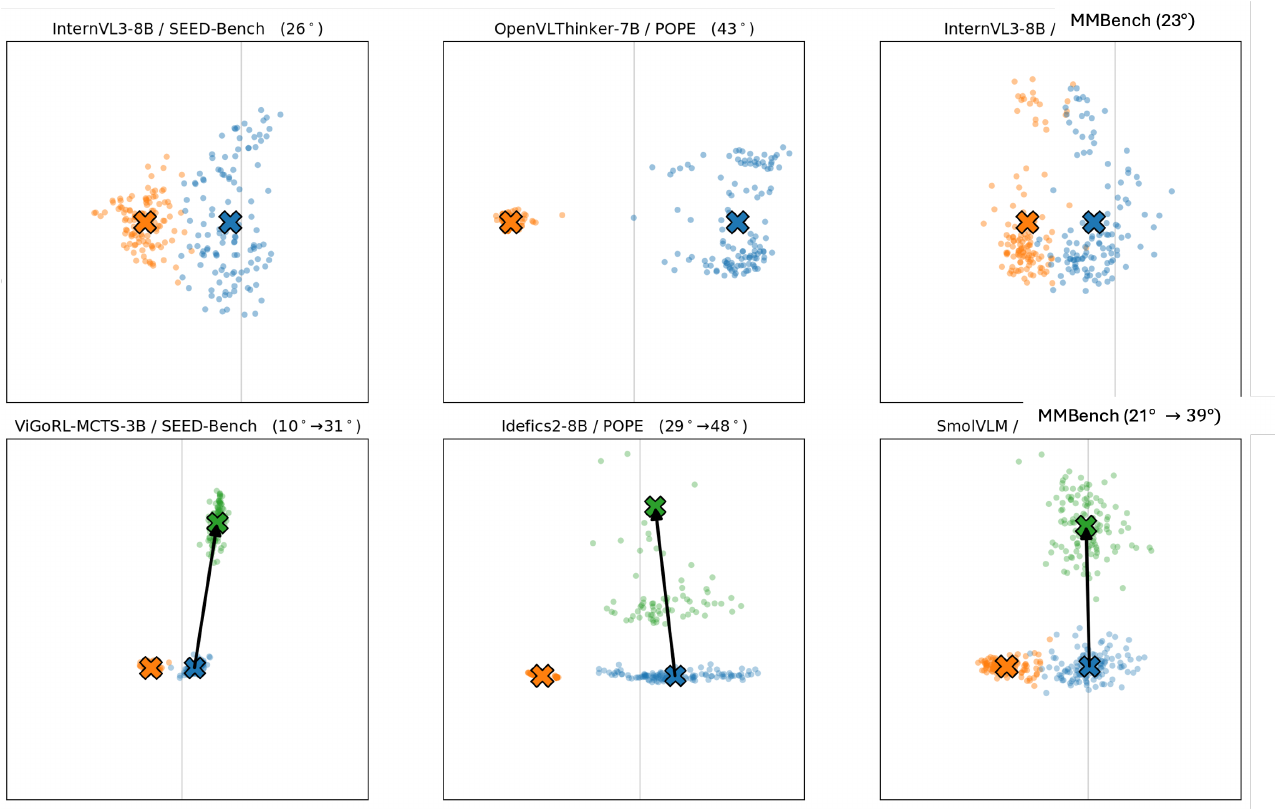}
\caption{Decision-state clouds. Top: Examples of weak text-dominant model-benchmark pairs, where the text-only (\textcolor{orange}{orange}) and text+image (\textcolor{blue}{blue}) decision states naturally maintain a distinct angular separation. Bottom: Examples of strong text-dominant pairs. Initially, their text+image states exhibited small angular differences from the text-only states, indicating heavy text reliance. However, our \textsc{GeoPID} vision-unique amplification effectively pushes the intervened states (\textcolor{green}{green}) away, establishing a clear separation. Crosses mark cloud centroids and arrows mark the before–after displacement. Displayed angular summaries are approximately $10^{\circ}\!\to31^{\circ}$ for ViGoRL-MCTS-3B~\citep{sarch2026grounded}/SEED-Bench~\citep{li2023seed}, $29^{\circ}\!\to48^{\circ}$ for Idefics2-8B~\citep{laurenccon2024matters}/POPE~\citep{li2023evaluating}, and $21^{\circ}\!\to39^{\circ}$ for SmolVLM~\citep{marafioti2025smolvlm}/MMBench~\citep{liu2024mmbench}. Reported angles are averaged over individual examples.}
\label{fig:f-clouds}
\end{figure}

\subsection{Decision-State Changes After Intervention}
\label{app:f-state-change}

\paragraph{Visualizing the image-conditioned displacement.}
Figure~\ref{fig:f-clouds} shows the decision-state clouds for three benchmark comparisons. \textcolor{orange}{Orange} points represent text-only states, \textcolor{blue}{blue} points the original image-and-text states, and \textcolor{green}{green} points the intervened states in the lower row. The intervened clouds move away from their original locations, and the displayed image-text angular separation increases in all three examples. These examples illustrate a representational change after the intervention. The displayed frame is constructed using an image-conditioned mean displacement, indicated in the asset as $w=\overline{H^{VT}}-\overline{H^T}$.


\section{More Ablation Studies}
\label{supp:ablations}
\providecommand{\GRobustFigsPath}{figure}
\providecommand{\GRobustFigurePath}{figure}
\subsection{Ablation Studies on Alternative Subspace Targets}
\label{app:f-specificity}

Table~\ref{tab:f-specificity} examines whether amplifying
$G_{U_v}$ yields greater gains than amplifying other
directions. We fix the normalized intervention depth
at $0.6$ and the strength at $\alpha=1$ across the same
$15$ model--benchmark pairs, changing only the amplified
subspace. In particular, the random text-complement
control tests whether orthogonality to the text subspace
alone is sufficient, or whether selecting the principal
directions of the residual vision representations matters.
As shown in Table~\ref{tab:f-specificity},
the vision-unique directions selected by \textsc{GeoPID}
yield the highest average accuracy gain.

\begin{table}[!htbp]
\centering
\small
\caption{
Alternative amplification targets on $15$ model-benchmark
pairs at normalized $\ell^*=0.6$ and $\alpha=1$.
Gain is in percentage points.
Correction and damage are reported percentages.
}
\label{tab:f-specificity}
\begin{tabular}{lrrrr}
\toprule
Amplified target & Rank & Gain & Correction & Damage \\
\midrule
Vision-unique $G_{U_v}$
& $32$ & $+0.47$ & $3.9$ & $1.4$ \\
Shared $G_R$
& --- & $-1.67$ & $4.1$ & $4.9$ \\
Text-unique $G_{U_t}$
& $7$ & $-0.47$ & $4.5$ & $3.2$ \\
Synergy-evaluation space $G_S$
& $51$ & $-0.27$ & $7.5$ & $4.6$ \\
Random hidden-space basis
& $32$ & $+0.07$ & $0.7$ & $0.3$ \\
Random text-complement basis
& $32$ & $+0.07$ & $0.6$ & $0.2$ \\
\bottomrule
\end{tabular}
\end{table}

\subsection{Ablation Studies on Different Intervention Settings}
\label{app:f-runtime-policy}

We examine an alternative rule for deciding which models
receive stronger intervention.
For each model, we compute the fraction of probe questions
whose predicted answers remain unchanged when the original
image is replaced with a blank image.
Models with an agreement rate of at least $0.7$ receive
stronger amplification.
Table~\ref{tab:f-policy-definition} specifies the full rule.

\begin{table}[!htbp]\centering\small
\caption{Full rule for alternative decision for which models receive stronger intervention.}
\label{tab:f-policy-definition}
\begin{tabular}{lp{.70\linewidth}}\toprule
Decision & Rule\\\midrule
Edit target & Vision-unique basis $G_{U_v}$ on visual tokens.\\
Open-branch strength & $\alpha=8$; or $\alpha=12$ when blank-image agreement $\geq0.9$.\\
Closed-branch strength & $\alpha=1$, a weak edit rather than the unedited baseline.\\
\bottomrule
\end{tabular}
\end{table}

\begin{figure}[!htbp]\centering
\includegraphics[width=\linewidth,trim=0bp 207bp 0bp 0bp,clip]{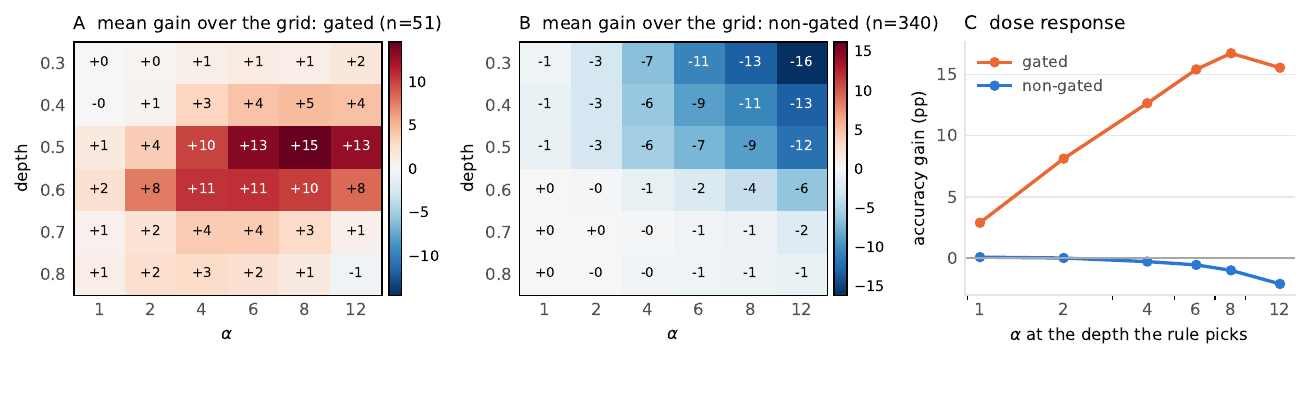}
\caption{Effects of intervention depth and amplification strength ($\alpha$) on accuracy for (A) gated ($n=51$) and (B) non-gated ($n=340$) cells, with (C) the corresponding performance trends across varying strengths at the selected depths.}
\label{fig:f-gate-dose}
\end{figure}

Figure~\ref{fig:f-gate-dose} shows how accuracy gains vary
with intervention depth $\ell^\star$ and strength $\alpha$.
The alternative policy selects a layer with the smallest
text-unique component within normalized depths $0.5$--$0.7$
and uses $\alpha=8$ for models meeting the agreement
threshold, while applying $\alpha=1$ to the remaining
models.
We also evaluate a variant using $\alpha=12$ when
agreement reaches $0.9$.
Table~\ref{tab:f-setting-comparison} compares the reported
mean gains with those of the rule specified in Figure~\ref{fig:rule}. It shows that the intervention rule in Figure~\ref{fig:rule} achieves a larger reported gain
($+4.62$ percentage points) than either alternative
($+2.24$ and $+2.28$ percentage points).



\begin{table}[!htbp]
\centering
\small
\caption{
Mean accuracy gains ($\%$) across $308$ model-benchmark pairs under different intervention selection rules. The primary result corresponds to Table~\ref{tab:percell-geotree_d4-filtered}.
}
\label{tab:f-setting-comparison}
\begin{tabular}{p{.76\linewidth}r}
\toprule
Intervention selection rule & Mean gain (pp) \\
\midrule
Blank-image agreement with geometry-based depth selection
& $+2.24$ \\
Blank-image agreement with $\alpha=12$ for agreement $\geq0.9$
& $+2.28$ \\
\textsc{GeoPID} geometry-based rule (Figure~\ref{fig:rule})
& $\mathbf{+4.62}$ \\
\bottomrule
\end{tabular}
\end{table}

\subsection{Ablation Studies on Different \texorpdfstring{$r$}{r}, \texorpdfstring{$\ell^*$}{ell}, and \texorpdfstring{$\alpha$}{alpha}}
\label{app:g-intervention-rank}

We vary the rank cap $r_u$ in Eq.~\ref{eq:unique_geometry}
and the intervention depth $\ell^\star$ and strength $\alpha$
in Eq.~\ref{eq:vision_intervention}.
Specifically, we evaluate $\alpha\in\{1,2,4\}$ at
normalized depths $\ell^\star/L\in\{0.5,0.6,0.7\}$
with $r_u=16$.
At depth $0.6$, we also evaluate $r_u=64$ using the
same three strengths.
Table~\ref{tab:g-rank-dose} reports mean accuracy gains
across POPE~\citep{li2023evaluating}, SEED-Bench~\citep{li2023seed}, and MMBench~\citep{liu2024mmbench}. With $r_u=16$, increasing $\alpha$ from $1$ to $4$
improves the mean gain at all three depths.
The largest gain among the tested settings is
$29.67$ percentage points at
$(r_u,\ell^\star/L,\alpha)=(16,0.6,4)$. At the same depth and strength, increasing $r_u$
to $64$ gives a smaller gain of $21.67$ points.
These results supplement the intervention settings
in Section~\ref{sec:geometry_intervention}.

\begin{table}[!htbp]\centering\small
\caption{Mean accuracy gains ($\%$) for ViGoRL-7B~\citep{sarch2026grounded} across
POPE~\citep{li2023evaluating}, SEED-Bench~\citep{li2023seed}, and MMBench~\citep{liu2024mmbench} under different rank caps,
normalized intervention depths, and strengths.}
\label{tab:g-rank-dose}
\begin{tabular}{rrrrr}\toprule
Rank cap & Depth & $\alpha=1$ & $\alpha=2$ & $\alpha=4$\\\midrule
$16$ & $0.50$ & $0.00$ & $+2.33$ & $+21.67$\\
$16$ & $0.60$ & $+4.33$ & $+19.33$ & $+29.67$\\
$16$ & $0.70$ & $+1.00$ & $+5.00$ & $+21.00$\\
$64$ & $0.60$ & $+3.67$ & $+10.33$ & $+21.67$\\\bottomrule
\end{tabular}
\end{table}


\section{Robustness}
\label{supp:robustness}
\subsection{Robustness to the Number of Image \texorpdfstring{$K$}{K} in Eq.~\ref{eq:synergy}}
\label{app:g-mismatches}

We vary the number of mismatched-image draws in
Eq.~\ref{eq:synergy} over $K\in\{3,5,10\}$
across $24$ model $\times$ benchmark pairs.
As shown in Table~\ref{tab:g-mismatch-k}, the reported
standard error of $S$ due to mismatch randomness
is approximately $0.001$,
indicating limited variability from the random
image assignments in this experiment.

\begin{table}[!htbp]\centering\small
\caption{Sensitivity to the number of mismatched-image draws
$K$ in Eq.~\ref{eq:synergy}.
Standard errors summarize mismatch-related variability
across $24$ model $\times$ benchmark pairs.}
\label{tab:g-mismatch-k}
\begin{tabular}{rrrr}\toprule
$K$ & Mismatch standard deviation & Reported SE of $S$ & Median $|S_K-S_1|$\\\midrule
$3$ & $0.0017$ & $0.0010$ & $0.0029$\\
$5$ & $0.0023$ & $0.0010$ & $0.0022$\\
$10$ & $0.0031$ & $0.0010$ & $0.0038$\\\bottomrule
\end{tabular}
\end{table}

\begin{table}[!htbp]\centering\small
\caption{Ablation study using $22$ models and $5$ datasets. Values are Spearman correlations between model rankings based on total decision-state information.}
\label{tab:g-estimator-ranking}
\begin{tabular}{llr}\toprule
Ablation & Setting & Rank correlation\\\midrule
Kernel & Cosine reference & $1.00$\\
 & RBF & $0.78$\\
 & Standardized linear & $0.75$\\\midrule
R\'enyi order & Shannon limit & $0.95$\\
 & $1.5$ & $1.00$\\
 & $2$ (reference) & $1.00$\\
 & $4$ & $1.00$\\\midrule
Sample fraction & $25\%$ & $0.99$\\
 & $50\%$ & $0.97$\\
 & $75\%$ & $1.00$\\
 & Full reference & $1.00$\\\bottomrule
\end{tabular}
\end{table}

\subsection{Robustness to Kernel, R\'enyi Order, and Sample Size}
\label{app:g-estimator}

We evaluate whether the model ordering by total
decision-state information in Eq.~\ref{eq:renyi_mi}
is preserved when changing the kernel, R\'enyi order,
and number of samples used in Section~\ref{sec:method}.
Using cached decision representations from $22$ models
and $5$ benchmarks, we recompute the information
measurements and compare the resulting model rankings
with the main-paper setting.

Section~\ref{sec:method} constructs the Gram matrix
$K_Z$ using cosine similarity:
\begin{equation}
K_Z[i,j]
=
\frac{\langle Z_i,Z_j\rangle}
{\|Z_i\|_2\|Z_j\|_2}.
\end{equation}
We replace this kernel with an RBF kernel,
\begin{equation}
K_Z^{\mathrm{RBF}}[i,j]
=
\exp\left(
-\frac{\|Z_i-Z_j\|_2^2}{2\sigma^2}
\right),
\end{equation}
where $\sigma$ is the bandwidth, or a standardized
linear kernel,
\begin{equation}
K_Z^{\mathrm{linear}}[i,j]
=
\langle \widetilde Z_i,\widetilde Z_j\rangle,
\end{equation}
where $\widetilde Z_i$ denotes the representation
after centering and scaling each feature across samples.
The RBF kernel assigns higher similarity to nearby
representations, while the standardized linear kernel
compares their inner products after standardization.
For each kernel, we apply the trace normalization
described in Section~\ref{sec:method} and recompute
total decision-state information using
Eq.~\ref{eq:renyi_mi}.

As shown in Table~\ref{tab:g-estimator-ranking},
the model rankings obtained with the RBF and
standardized linear kernels have Spearman correlations
of $0.78$ and $0.75$ with the cosine-kernel reference.
Thus, the rankings remain positively associated.
Keeping the cosine kernel and replacing the
R\'enyi-2 entropy in Eq.~\ref{eq:renyi_mi}
with orders $1.5$ and $4$ yields ranking correlations
of $0.996$ and $0.997$, respectively.
Reducing the sample count to $25\%$, $50\%$, or $75\%$
also preserves the ranking closely, with reported
correlations of approximately $0.99$, $0.97$, and
$1.00$.



\end{document}